\documentclass[letterpaper, 10 pt, conference]{ieeeconf}  % Comment this line out if you need a4paper

\IEEEoverridecommandlockouts                              % This command is only needed if 
\usepackage{bbm}
\usepackage{etoolbox}
\usepackage{multicol}
\usepackage[bookmarks=true]{hyperref}
\usepackage{graphics} % for pdf, bitmapped graphics files
\usepackage{times} % assumes new font selection scheme installed
\usepackage{amsmath} % assumes amsmath package installed
\usepackage{amssymb}  % assumes amsmath package installed
\usepackage{mathrsfs}
\usepackage{amsfonts}
\usepackage{kantlipsum}
\usepackage{subfigure}
\usepackage{marvosym}
\usepackage{graphicx}
\usepackage{float}
\usepackage{ctable}
\usepackage{cuted}
\usepackage{colortbl}
\usepackage{multirow}
\usepackage{multicol}
\usepackage{cite}
\usepackage{wrapfig}
\usepackage{algorithm}
\usepackage{algorithmic}
\usepackage[algo2e]{algorithm2e}

\DeclareMathAlphabet{\mathmybb}{U}{bbold}{m}{n}

\definecolor{Gray}{gray}{0.9}
\definecolor{White}{gray}{1}

\title{\LARGE \bf
ManiFoundation Model for General-Purpose Robotic Manipulation of Contact Synthesis with Arbitrary Objects and Robots}

\author{Zhixuan Xu$^{1*}$, Chongkai Gao$^{1*}$, Zixuan Liu$^{2*}$, Gang Yang$^{1*}$, Chenrui Tie$^{3}$, Haozhuo Zheng$^{4}$, Haoyu Zhou$^{5}$,\\ Weikun Peng$^{1}$, Debang Wang$^{1}$, Tianrun Hu$^{1}$, Tianyi Chen$^{6}$, Zhouliang Yu$^{7}$, Lin Shao$^{1}$\textsuperscript{\textdagger}
{
\thanks{* denotes equal contribution}
\thanks{\textdagger \ denotes the corresponding author}
\thanks{$^{1}$Zhixuan Xu, Chongkai Gao, Gang Yang, Weikun Peng, Debang Wang, Tianrun Hu, and Lin Shao are with the Department of Computer Science, National University of Singapore.
        {\tt\small ariszxxu@gmail.com, gaochongkai@u.nus.edu, yg.matinal@gmail.com,
        debang@u.nus.edu, tianrunhu@gmail.com, linshao@nus.edu.sg}}%
\thanks{$^{2}$Zixuan Liu is with Tsinghua Shenzhen International Graduate School, Tsinghua University.
        {\tt\small zx-liu21@mails.tsinghua.edu.cn}}%
\thanks{$^{3}$Chenrui Tie is with School of Electronics Engineering and Computer Science, Peking University. 
{\tt\small crtie@pku.edu.cn}}
\thanks{$^{4}$Zheng Haozhuo is with Department of Mathematics, National University of Singapore.
          {\tt\small muztaga2@gmail.com}}%
\thanks{$^{5}$Haoyu Zhou is with Department of Mechanical Engineering, National University of Singapore
        {\tt\small zhouhaoyu01@u.nus.edu}}
\thanks{$^{6}$Tianyi Chen is with the Department of Computer Science and Engineering, Shanghai Jiao Tong University
        {\tt\small tianyyiii@gmail.com}}
\thanks{$^{7}$Zhouliang Yu is with the Division of Emerging Interdisciplinary Areas, Hongkong University of Science and Technology
        {\tt\small zhouliangyu@link.cuhk.edu.cn}}
}
}

\begin{document}

\maketitle
\thispagestyle{empty}
\pagestyle{empty}

\begin{abstract}
To substantially enhance robot intelligence, there is a pressing need to develop a large model that enables general-purpose robots to proficiently undertake a broad spectrum of manipulation tasks, akin to the versatile task-planning ability exhibited by LLMs. The vast diversity in objects, robots, and manipulation tasks presents huge challenges. Our work introduces a comprehensive framework to develop a foundation model for general robotic manipulation that formalizes a manipulation task as contact synthesis. Specifically, our model takes as input object and robot manipulator point clouds, object physical attributes, target motions, and manipulation region masks. It outputs contact points on the object and associated contact forces or post-contact motions for robots to achieve the desired manipulation task. We perform extensive experiments both in the simulation and real-world settings, manipulating articulated rigid objects, rigid objects, and deformable objects that vary in dimensionality, ranging from one-dimensional objects like ropes to two-dimensional objects like cloth and extending to three-dimensional objects such as plasticine. Our model achieves average success rates of around 90\%. Supplementary materials and videos are available on our project website at~\href{https://manifoundationmodel.github.io/}{https://manifoundationmodel.github.io/}. 
\end{abstract}

\section{Introduction}
\vspace{-1mm}
Foundation models~\cite{bommasani2021opportunities} trained on broad data are becoming versatile tools for numerous tasks, revolutionizing fields like natural language processing and computer vision. Large Language Models~(LLMs) and Vision-Language Models~(VLMs) serve as vast repositories of knowledge that can enhance robotic capabilities. For instance, LLMs have demonstrated superior capacity in decomposing complex tasks, such as meal preparation, into subtasks, significantly improving robot planning and reasoning processes. Recently, researchers have started investigating LLMs and VLMs in robotics to develop robot foundation models. These efforts either leverage pretrained LLMs or VLMs as high-level planners to execute a set of pretrained low-level skills~\cite{ahn2022can,huang2023voxposer,codeaspolicies2022}, or train low-level robot actions on massive datasets~\cite{brohan2022rt,brohan2023rt,padalkar2023open}. While these models excel at high-level task planning and semantic understanding, they face challenges with the complexity of 3D object geometry, deformations, force constraints, and comprehending different robot manipulator topologies during low-level manipulations. In other words, the breadth and adaptability of low-level manipulation capabilities have received less attention compared to high-level task planning in robotics. Thus, it is essential to build a \textit{manipulation foundational model} that equips robots with the requisite low-level manipulation proficiency to engage with the physical world and competently execute a diverse range of tasks.

\begin{figure}[t] \centering
    \includegraphics[width=0.95 \linewidth]{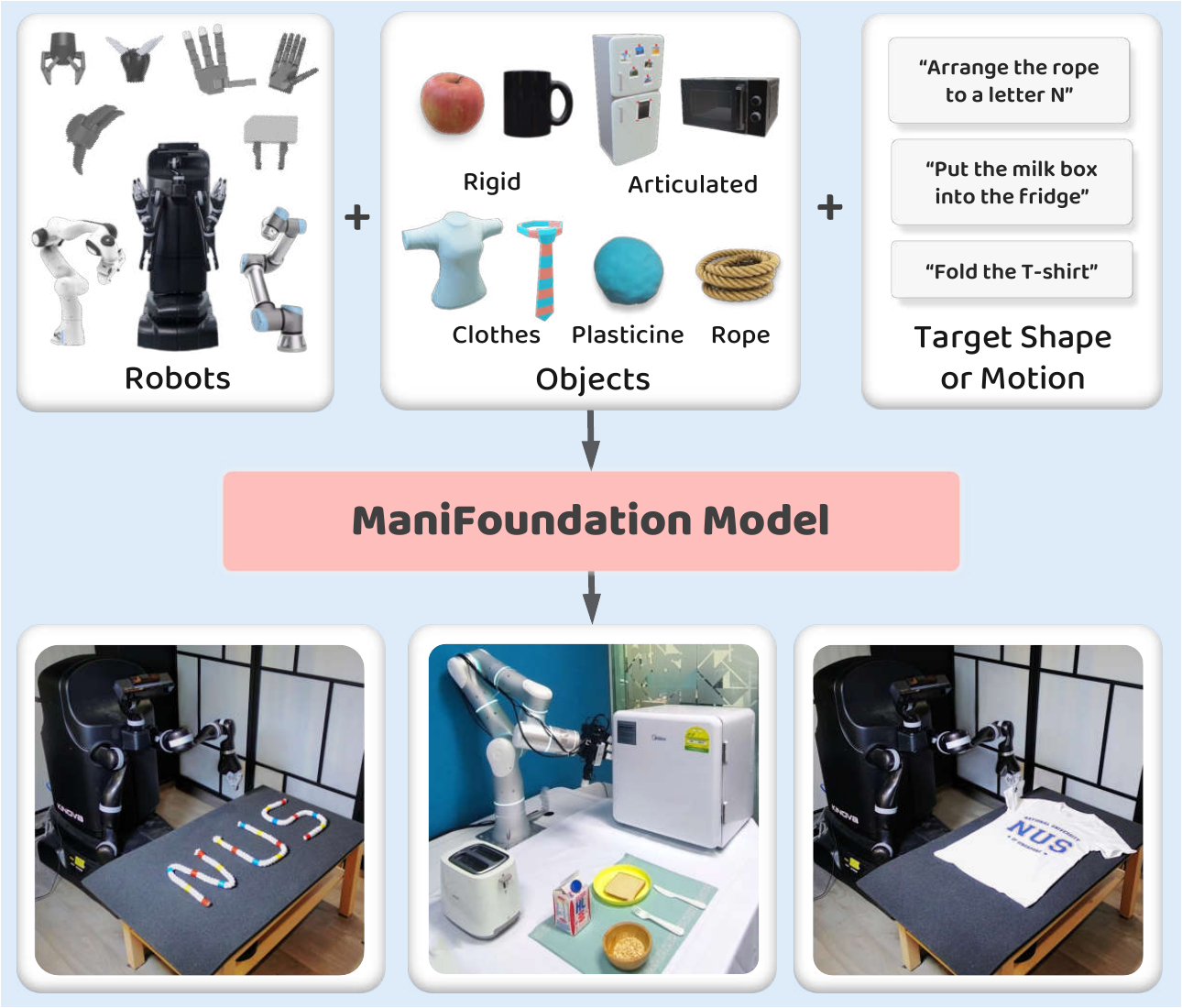}
    \vspace{-5pt}
    \caption{We propose a ManiFoundation Model that can generalize over a diverse range of robots and objects, and perform various kinds of manipulation tasks based on 3D point cloud input.} \label{fig:figure1}
    \vspace{-15pt}
\end{figure}

To build a manipulation foundation model, the core problem is defining a unified task formulation for all robot manipulation tasks, similar to auto-regressive prediction in GPTs~\cite{gpt}, mask prediction in BERT~\cite{bert}, and vision-language contrastive learning in CLIP~\cite{clip}. This formulation must: 1) handle arbitrary objects, including articulated rigid objects, solid rigid objects, and deformable objects; 2) support a diverse range of manipulation tasks, such as grasping, pulling, re-orientating, and folding; 3) adapt to cross-platform integration, accommodating various robot designs and configurations; and 4) integrate seamlessly with LLMs/LVMs for effective task execution and long-horizon planning.

In this work, we propose a foundation model framework for low-level robotic manipulation to meet the above requirements. We formulate the manipulation foundation task as a contact synthesis task. Specifically, the foundation model $\mathcal{F}$ takes as input: 1) object point cloud and physical attributes (e.g., friction coefficients, Young’s modulus), 2) task descriptions as target motion, 3) robot manipulator descriptions as point clouds, and 4) manipulation region masks if required (e.g., grasping a pan's handle). The foundation model $\mathcal{F}$ outputs the contact points, contact forces, or post-contact motion to enable the robot to manipulate the object towards its target motion. We generated a large-scale annotated dataset to train the model. We tested our ManiFoundation model in diverse experiments, verifying its effectiveness in both simulations and real-world applications. In summary, our primary contributions are:
\begin{itemize}
    \item We propose a comprehensive framework for foundation models in general-purpose robotic manipulation via contact synthesis, achieving broad generalization across various objects, robots, and tasks.
    \item We implement the ManiFoundation model, which includes neural network backbones for visual and physical feature extraction and an iterative contact wrench optimization method to generate feasible robot poses while avoiding collisions and penetrations.
    \item We develop a pipeline to generate a large-scale annotated dataset covering diverse objects and robot hands, which will be publicly available on our project website.
    \item We conduct extensive experiments to verify the effectiveness of our framework in both simulations and real-world settings, covering various articulated rigid and deformable object manipulation tasks.
\end{itemize}
%===============================================================================

\section{Related Work}
%We review related literature on key components in our approach, including foundation model framework for robotics, learning visual affordances, contact synthesis optimization, and deformable object manipulation. We describe how we are different from previous work.W
% We review related literature on two key components in our approach and provide a comprehensive review in the supplementary material.
\subsection{Foundation Models in Robotics}
Foundation models in robotics~\cite{firoozi2023foundation} have recently garnered significant attention. SayCan~\cite{ahn2022can} grounds LLMs using value functions of pretrained skills. VoxPoser~\cite{huang2023voxposer} combines affordances and constraints from LLMs and vision-language models into 3D value maps for motion planners. Codes-as-policies~\cite{codeaspolicies2022} leverages LLMs to write robot policy code from natural language instructions. Robotics Transformer~(RT-1)~\cite{brohan2022rt} was trained on 130k real-world data points, covering over 700 tasks collected by 13 robots over 17 months. RT-2\cite{brohan2023rt} enhances vision-language-action models using web and robotic data. RT-X~\cite{padalkar2023open} collects a dataset from 22 robots with 527 robotic skills, showing that scaling data improves generalization across multiple robot platforms. Our work focuses on developing a foundation model specifically designed for manipulation skills via contact synthesis, supporting general robotic manipulation with broad generalization across diverse deformable objects, robots, and tasks.

\subsection{Learning Visual Affordances}
Learning affordances~\cite{do2018affordancenet,yen2020learning,pmlr-v205-lin23c} has received increasing attention for robotic manipulation, including grasping~\cite{mandikal2021learning,borja2022affordance,wu2023learning} and articulated object manipulation~\cite{Mo_2021_ICCV,wu2021vat,wang2022adaafford}. Where2Act~\cite{Mo_2021_ICCV} predicts the actionable point-on-point cloud for primitive actions such as pushing and pulling. Vision-Robotic Bridge~(VRB)~\cite{bahl2023affordances} train a visual affordance model on large-scale videos with human behavior to predict contact points and a post-contact trajectory learned from human videos. Our model provides contact points for diverse robot hands and objects, including dexterous hands and deformable objects, together with contact forces and post-contact motions to guide robot low-level action execution.

\section{ManiFoundation Model Design}
\begin{figure*}[t]
\centering
\includegraphics[width=0.98\textwidth]{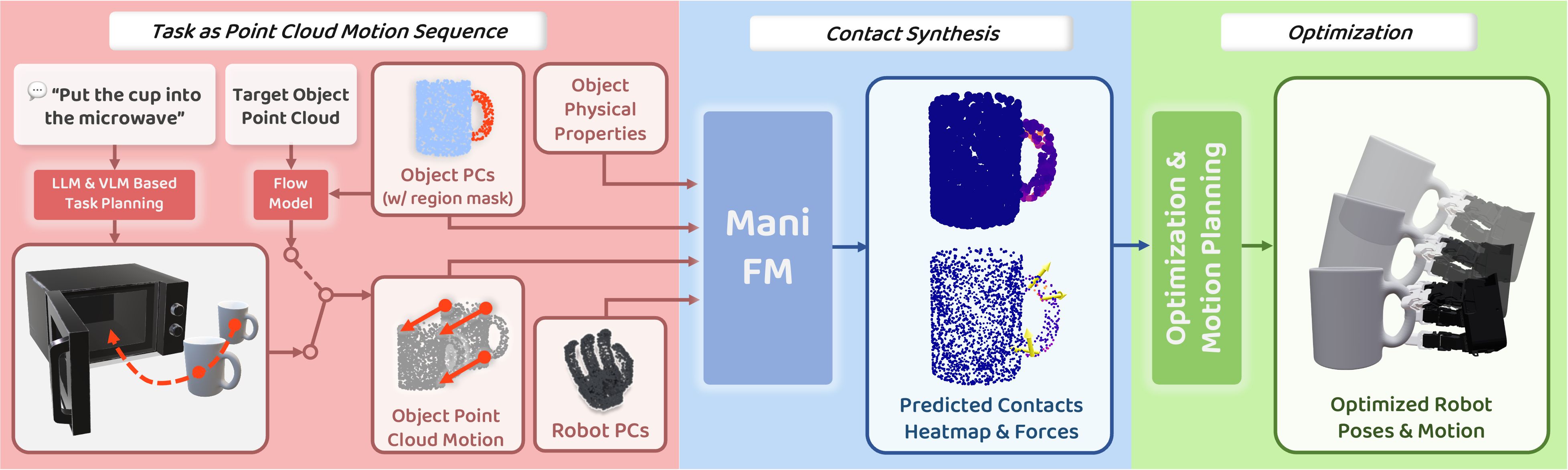}
\vspace{-5pt}
\caption{The pipeline of our ManiFoundation model. Left: we decompose a manipulation task to a sequence of object point cloud motions from either VLM-based planning or a flow model. Middle: we train a ManiFoundation network to predict the contact point and force heatmap for each motion of the sequence. Right: we acquire the robot pose for execution based on optimization with the initial results from the contact point and force heatmaps.}
\label{fig:overview}
\vspace{-15pt}
\end{figure*}

\begin{figure}[t]
\centering
\vspace{5pt}
\includegraphics[width=0.49\textwidth]{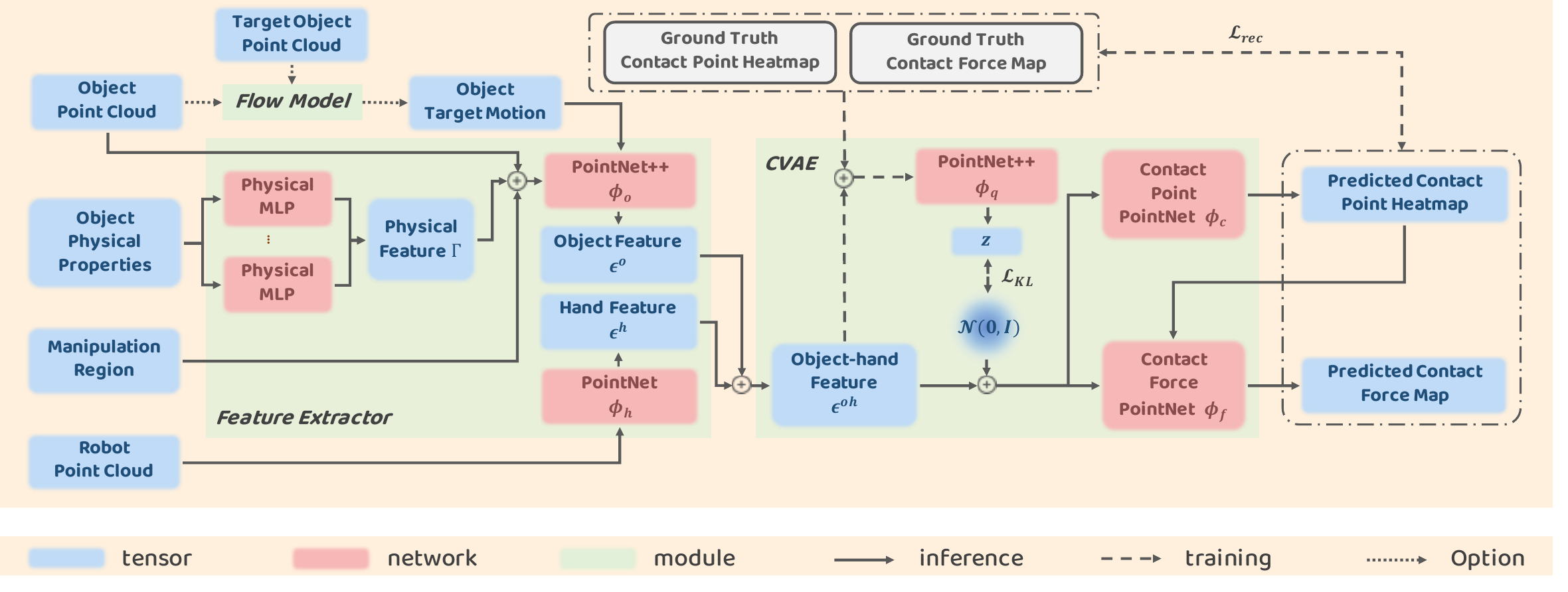}
\vspace{-20pt}
\caption{Overview of our ManiFoundation network. The feature extractor module incorporates the information from both object and robot point clouds, and the CVAE module generates the contact point and force maps on the given object.}
\label{fig:network}
\vspace{-15pt}
\end{figure}

A robotic manipulation task can be divided into two parts: a task-planning module that determines the object trajectory, and a manipulation module that guides the robot to achieve this trajectory. For any given task, we assume LLMs/VLMs, other models, or differentiable simulators with digital twins can convert it into a sequence of object point clouds as subgoals. Examples of this step for tasks in our experiments are provided on our website. Note that this aspect is not our primary focus. Here, we aim to develop a universal model that guides robots to manipulate objects towards the target pose or shape. The model should generalize across arbitrary objects, robots, and tasks (target shape or motion).

We present our ManiFoundation model as a solution to the \textit{contact synthesis} problem. Specifically, the model $\mathcal{F}$ takes as input: 1) manipulated object descriptions, including the object point cloud and physical properties; 2) task descriptions as target motion; 3) robot descriptions as point clouds; and 4) manipulation region masks if needed (e.g., a pan's handle). The model outputs contact points and contact forces to manipulate the object towards its target motion. Formally, let a point cloud with $N$ points and normals on each point be $\mathcal{P}={(x_1, n_1),\cdots,(x_N,n_N)} \in \mathbb{R}^{N\times 6}$, where $x_i\in \mathbb{R}^3$ is the position and $n_i\in \mathbb{R}^3$ is the inward normal vector of the $i$-th point. Let $\mathcal{P}o$ and $\mathcal{P}h$ be the input object and robot hand point clouds, respectively. Let $\mathcal{M}=\{m_i\}_{i=1}^{N}, m_i\in \mathbb{R}^3$ be the task motion that should be achieved on each point of the object. Let the physical feature of the object be $\Gamma$ and the region mask be $\mathcal{R}$. The manipulation foundation problem is defined as finding a contact solution $M_{\mathcal{C}}, M_{\mathcal{S}} = \mathcal{F}(\mathcal{P}_o, \mathcal{P}_h, \mathcal{M}, \mathcal{R}, \Gamma)$ that satisfies: \vspace{-5pt}
\begin{equation} 
M_{\mathcal{C}}, M_{\mathcal{S}} = \min \|\mathbf{f}(\mathcal{P}_o, \mathcal{P}_h, \Gamma, M_{\mathcal{C}}, M_{\mathcal{S}}) - \mathcal{M}\|
\end{equation}

\vspace{-5pt} where $M_{\mathcal{C}}=\{c_i\}_{i=1}^{H}, c_i \in \mathbb{R}^3$ is the predicted contact points with $H$ points on the object point cloud, and $M_{\mathcal{S}}=\{s_i\}_{i=1}^{H}, s_i \in \mathbb{R}^3$ is the contact force corresponds to each contact point $c_i$. $\mathbf{f}$ stands for the forward dynamics function that outputs the object per-point motion given the contact points and forces. Figure~\ref{fig:overview} provides a pipeline overview. The following sections describe each module in detail.

%As introduced above, the role of the neural network in our method is to predict $\mathcal{C}, \mathcal{S}$ as the initial solutions for the downstream optimization process. The strong function approximation ability of neural networks can deal with the diverse kinds of input objects, hands, and task motions, which can reduce the burden of the optimization process and increase success rates. However, there are several main challenges here, which are: 1) different kinds of physical properties among rigid body, cloth, and partial objects and various geometric shapes; 2) the predicted contact points are kinematically feasible; 3) multiple solutions for the same given task motion. We design specific modules for these challenges. 

\subsection{The ManiFoundation Network}~\label{sec:NN_arch}
The whole network structure is illustrated in Figure \ref{fig:network}. There are mainly two parts of our network: the Feature Extractor and the Conditional Variational Autoencoder (CVAE). The feature extractor is designed to encode information from the object and robot manipulator, while the CVAE module is designed to fuse extracted features to generate contact points and force heatmaps for execution. We introduce the details of each module in the following sections.

\subsubsection{Feature Extractor} 
% Let a point cloud with $N$ points and normals on each point be $\mathcal{P}=\{(x_1, n_1),\cdots,(x_N,n_N)\} \in \mathbb{R}^{N\times 6}$, where $x_i\in \mathbb{R}^3$ is the position and $n_i\in \mathbb{R}^3$ is the inward normal vector of the $i$-th point. Let $\mathcal{P}_o$ and $\mathcal{P}_h$ be the input object and robot hand point cloud respectively. Let $\mathcal{M}=\{m_i\}_{i=1}^{N}, m_i\in \mathbb{R}^3$ be the task motion that should be achieved on each point of the object. 
We encode the object point cloud and robot manipulator point cloud separately, using \textit{PointNet++}\cite{qi2017pointnet++} ($\phi_o$) for $\mathcal{P}_o$ and \textit{PointNet}\cite{pointnet} ($\phi_h$) for $\mathcal{P}_h$. We chose \textit{PointNet} for the manipulators due to its simpler architecture and fewer parameters, making it suitable for the limited variety of manipulators. In contrast, \textit{PointNet++} is better suited for extracting features from the more diverse and complex object point clouds. To address the diversity in both objects and task requirements, we also design specific modules before these feature extractors.

\paragraph{Physical Properties Encoder}
Our model accommodates a wide variety of objects, each with distinct physical properties that are crucial for manipulation. These properties dictate different manipulation strategies. For example, a handkerchief and a card of the same shape require different approaches: spinning a handkerchief involves grasping the four corners, while rotating a card involves applying friction to other areas. These differences necessitate encoding physical attributes to distinguish between different types of objects, leading to varied contact points and force solutions.

In this work, we encode different physical properties with different Multi-layer Perceptions (MLPs) $\psi_i: \mathbb{R}^1 \rightarrow \mathbb{R}^{10}, i=1,\cdots,6$. The same MLP processes the same physical properties across different points. Specifically, we encode friction coefficients for all objects, density and elasticity coefficients~\cite{li2022diffcloth} for 2D deformable objects, and Young's modulus and Poission's ratio~\cite{liu2023softmac} for 1D and 3D deformable objects. Each physical property is a one-dimensional value on each point of $\mathcal{P}_o$. All physical features extracted by separate MLPs are aggregated using Masked Mean Pooling (MMP), where the missing physical properties on certain types of objects are masked. Thus, for each point of $\mathcal{P}_o$, the extracted physical feature $\Gamma$ is $\Gamma = MMP(\psi_1(\gamma_1), \cdots, \psi_6(\gamma_6)) \in \mathbb{R}^{10}$,
where $\gamma_i$ is the $i$-th physical property. 

This design offers several benefits. First, it can handle objects with a wide range of physical properties. Second, if additional properties are needed to identify new object types, new MLPs can be easily added to map these properties to feature vectors. These new vectors are then combined with existing ones and fed into the mean pooling layer.

\paragraph{Manipulation Region Selection}~\label{sec:contactSelection}
Certain manipulation tasks have constraints that specify which parts of an object can be manipulated, often due to human preferences or environmental factors. For example, robots are typically preferred to grasp the handle of a teapot or pan and cannot grasp the bottom of a cup when it's on a table. To address these constraints, we design a manipulation region mask, denoted as $\mathcal{R} =\left\{{r_i}\right\}_{i=1}^{N} \in \mathbb{R}^{N}$. $r_i = 1$. Here, $r_i = 1$ indicates that the $i$-th point of $\mathcal{P}_o$ is allowed to contact, while $r_i = 0$ means it is not. This mask can be assigned by humans or generated based on environmental constraints.
% \paragraph{Optional Target Motion Prediction} 

For deformable objects, specifying point-wise target motion is challenging. We use a scene flow model \cite{wang2022matters} to process both current and target point clouds to deduce the target motion. For articulated or rigid objects, determining the target motion is straightforward.

In summary, $\phi_o$ takes as inputs object point cloud $\mathcal{P}_o$, the task motion $\mathcal{M}$, the physical feature $\Gamma$, and the manipulation region mask $\mathcal{R}$, and output the extracted feature vector for each point of the object. $\phi_h$ take as input only the hand point cloud $\mathcal{P}_h$. That is:\vspace{-3pt}
\begin{equation}
    \begin{split}
        \epsilon^o_i & = \phi_o(x_i, n_i, m_i, r_i, \Gamma), i=1,\cdots,N_o. \\
        \epsilon^h_i & = \phi_h(x_i, n_i), i=1,\cdots,N_h.
    \end{split}
\end{equation}

\vspace{-3pt} We then perform max pooling on the extracted hand features to get a global hand feature $\epsilon^h$ and concatenate each object point feature $\epsilon_i^o$ with $\epsilon^h$ to get a combined feature point cloud $\epsilon^{oh}=\{\epsilon_{i}^{oh}\}, i=1,\cdots, N_o+N_h$. Next, the following CVAE will fuse them to get the final output.

\subsubsection{CVAE for Contact Point and Force/Motion}
In practice, multiple contact points and force combinations can achieve the same target motion for an object. For instance, to advance a cube, one might push from behind or pull from its sides. To generate a range of solutions, we utilize a Conditional Variational Autoencoder (CVAE) for creating contact point heatmap and force/motion heatmap. The contact point heatmap, with dimensions $\mathbb{R}^{N\times 1}$, assigns a probability score to each point indicating its likelihood of being a contact point. The force/motion map, sized $\mathbb{R}^{N\times 3}$, represents either the contact force or the post-contact motion at each point. We chose CVAE instead of diffusion models for its simpler structure, faster inference speed, and multimodal generation capabilities as demonstrated in our experiments.

Specifically, during training, the ground truth contact point heatmap $M_{\mathcal{C}'}$ and force/motion map $M_{\mathcal{S}'}$ together with $\epsilon^{oh}$ are encoded into latent variables $z\in\mathbb{R}^{64}$ by a posterior \textit{PointNet++} $\phi_q$: $z = \phi_q (\epsilon^{oh},M_{\mathcal{C}'}, M_{\mathcal{S}'})$, and $z$ will be concatenated on each point of the condition feature $\epsilon^{oh}$ to be sent to following decoders.  During the inference phase, the prior latent variables are sampled from a Gaussian distribution. To get the contact points and force/motion maps~\label{sec:contactmap}, we use two \textit{PointNet}s $\phi_c$ and $\phi_f$ to predict the contact point heatmap $M_\mathcal{C}$ and contact force/motion map. We then use non-maximum suppression (NMS)~\cite{neubeck2006efficient} to select $H$ contact points $\mathcal{C}$, where $H$ is less than or equal to the finger number of the hand. With the predicted contact heatmap $M_\mathcal{C}$ as an extra input, we predict the force map $M_\mathcal{S}$ of shape $\mathbb{R}^{N\times 3}$, composed of 3D vectors on each point of the object by $\phi_f$. We select the forces/motions on the chosen contact points as the final contact force/motion $\mathcal{S}$.

The training objectives of the whole network include the mean-square loss between the predicted and ground-truth contact point and force heatmaps, as well as the CVAE training loss, that is: \vspace{-3pt}
\begin{equation}
    \begin{split}
    \mathcal{L}&=\lambda_{\mathcal{C}}\mathcal{L}_{MSE}(M_{\mathcal{C}'}, M_{\mathcal{C}}) + \lambda_{\mathcal{S}}\mathcal{L}_{MSE}(M_{\mathcal{S}'}, M_{\mathcal{S}}) \\ &+ \lambda_{KL}\mathcal{D}_{KL}(\phi_q (\epsilon_{oh},M_{\mathcal{C}'}, M_{\mathcal{S}'}) \mid\mid \mathcal{N}(0,I)),
    \end{split}
\end{equation}

 \vspace{-3pt} where
$\lambda_{\mathcal{C}}, \lambda_{\mathcal{S}}, \lambda_{KL}$ are hyperparameters for loss weights. $\mathcal{N}(0,I)$ is a standard Guassian distribution.% we assmue the latent variables to have.

% \subsection{ManiFoundation Model: Contact Wrench Generation}
\subsection{Robotic Hand Pose Refinement}~\label{sec:opt_refine}
Given the contact points predicted by our network, we can directly manipulate deformable objects by simply grasping the contact points to achieve the target motion. However, articulated and rigid object manipulation generally requires the coordinated effort of all fingers, while the contact points predicted by our network may not lead to feasible joint values of the robotic hand that satisfy the robot kinematics and avoid penetrations between robots and objects. Thus, we develop an optimization module for robotic hand pose refinement. 

\subsubsection{Pose Initialization}
We infer the initial hand pose from the predicted contact points described in Sec.\ref{sec:NN_arch}. For a multi-fingered hand, we iterate all possible matches between fingertips and contact points (e.g., $4!=24$ matches for a four-fingered hand). For each match, we fix the robot hand in its neutral joint values as a rigid object and apply the Iterative Closest Point (ICP) algorithm to align the fingertips with the predicted contact points. The resulting hand 6D pose for each match is used to generate a contact heatmap, and the best match, based on the highest Intersection over Union (IoU) score between the generated and predicted heatmaps, is selected. This initial alignment provides a good starting pose for the robot palm, roughly aligning fingertip positions with predicted contact points, thereby shortening subsequent optimization time and improving success rates.

\subsubsection{Target Motion and Target Wrench}
Given the target motion of the rigid object, our system uses inverse dynamics to infer the target wrench $\boldsymbol{w}$ for producing the specified motion. Specifically, the target motion over time interval $\Delta t$ involves a rotation $\Delta \boldsymbol{\theta}$ followed by a translation $\Delta \boldsymbol{x}$, with $\Delta \boldsymbol{\theta}$ indicating a rotation around the axis $\frac{\Delta \boldsymbol{\theta}}{\vert \Delta \boldsymbol{\theta} \vert}$ through an angle of $\vert \Delta \boldsymbol{\theta} \vert$ radians. It can be calculated as follows: \vspace{-10pt}

{\footnotesize
\begin{align}
     & \boldsymbol{\hat{w}} = \frac{2}{\Delta t^2} 
    \begin{bmatrix}
        \Delta \boldsymbol{x} / M \\
        \Delta \boldsymbol{\theta} / I
    \end{bmatrix}, \\
    & \boldsymbol{w}= \boldsymbol{\hat{w}} / \|\boldsymbol{\hat{w}}\|,
\end{align}}

\vspace{-6pt} where object mass and inertia are denoted as $M$ and $I$, which are computed approximately by estimating the volume of the object. %\lin{A default value like $M=1kg, I=0.002\mathbf{I}_3$ also works well in our experiments.} 
We normalize $\boldsymbol{\hat{w}}$ to target wrench $\boldsymbol{w}$ for simplicity. In practice, we care more about the direction of the wrench and leave low-level force/torque control to take care of action to move objects towards target shapes.%and multiply it with a coefficient $\lambda_{\boldsymbol{w}}=0.2$. Since all the contact forces have a max magnitude of one in their normal direction during numerical optimization, a coefficient $\lambda_{\boldsymbol{w}}$ less than one is equivalent to relaxing the force magnitude limit and increasing the search space. \lin{why we need to have a coeff?}. 

\subsubsection{Wrench Optimization}
\label{sec:opt}
Given a target wrench and robot hand's joint poses, we calculate the optimal contact forces $\boldsymbol{f}^*$ exerted by $m$ fingertips onto the object to minimize the norm of wrench residual $\boldsymbol{r}$ in Eqn.~\ref{eqn:wrenches}. 

{\footnotesize
\begin{align}
     \boldsymbol{r}(\boldsymbol{w}, \boldsymbol{q}, \boldsymbol{f}_{1:m}) & = \sum_{i=1}^m G_i(\boldsymbol{q}) \boldsymbol{f}_i - \boldsymbol{w}, \label{eqn:wrenches}\\ 
    \boldsymbol{f}_{1:m}^*(\boldsymbol{w}, \boldsymbol{q}) & = \underset{\boldsymbol{f}_{1:m}}{\arg\min} \;  \Big\Vert \boldsymbol{r}(\boldsymbol{w}, \boldsymbol{q}, \boldsymbol{f}_{1:m}) \Big\Vert ^2, \; \\ \textrm{s.t. } \boldsymbol{f}_{1:m} & \in FC(\mu).
\end{align} }

\vspace{-8pt} where $G_{1:m}$ is the grasp mapping matrices that maps 3d contact force at fingertips to 6d wrench applied to the object. $FC(\mu)$ is the friction cone with coefficient $\mu$ guaranteeing that contact forces satisfy Coulomb's law~\cite{boyd2007fast}. During optimization, fingertips are not required to always be in contact with the object surface. If fingertip $i$ is away from the surface, $G_i$ is calculated from its position $\boldsymbol{p}_i$ and the normal vector at its projection onto the surface $\bar{\boldsymbol{n}}_i$.

The optimization problem to improve grasp quality is: \vspace{-12pt}

% {\footnotesize
% \begin{align}
%     \min_{\boldsymbol{q}} \; &  \Big\Vert \boldsymbol{r}^*(\boldsymbol{w}, \boldsymbol{q}) \Big\Vert ^2  + \sum_{i=1}^m \Big\vert \Phi \big(\boldsymbol{p}_i(\boldsymbol{q}) \big) \Big\vert ^2, \label{eqn:objective} \\
%     \textrm{s.t. }& \boldsymbol{q} \in [\boldsymbol{q}_{min}, \boldsymbol{q}_{max}], \label{eqn:c_limit} \\
%     &\Phi \big(\boldsymbol{c}_{1:n}(\boldsymbol{q}) \big) \ge \epsilon_{1:n}, \label{eqn:c_penetration} \\
%     &\Big\Vert \boldsymbol{c}_i(\boldsymbol{q}) - \boldsymbol{c}_j(\boldsymbol{q}) \Big\Vert \ge \epsilon_i + \epsilon_j, \label{eqn:c_self_penetration}
% \end{align}
% }
\noindent
{
\footnotesize
\begin{align}
    \min_{\boldsymbol{q}} \; & \Big\Vert \boldsymbol{r}^*(\boldsymbol{w}, \boldsymbol{q}) \Big\Vert ^2  + \sum_{i=1}^m \Big\vert \Phi \big(\boldsymbol{p}_i(\boldsymbol{q}) \big) \Big\vert ^2, \label{eqn:objective} \\
    \textrm{s.t. }& \boldsymbol{q} \in [\boldsymbol{q}_{min}, \boldsymbol{q}_{max}], \label{eqn:c_limit} \\
    & \Phi \big(\boldsymbol{c}_{1:n}(\boldsymbol{q}) \big) \ge \epsilon_{1:n}, \label{eqn:c_penetration} \\
    & \Big\Vert \boldsymbol{c}_i(\boldsymbol{q}) - \boldsymbol{c}_j(\boldsymbol{q}) \Big\Vert \ge \epsilon_i + \epsilon_j, \label{eqn:c_self_penetration}
\end{align}
}

\vspace{-5pt} where $\boldsymbol{r}^*(\boldsymbol{w}, \boldsymbol{q}) = \boldsymbol{r} \big(\boldsymbol{w}, \boldsymbol{q}, \boldsymbol{f}_{1:m}^*(\boldsymbol{w}, \boldsymbol{q}) \big)$ is the minimum wrench residual. Our optimization objective in Eqn.~\ref{eqn:objective} comprises two terms. The first term is the minimized norm of wrench residual, which measures the current grasp's capability to generate the desired object motion. The second term is the sum of the distance between fingertips and the object's surface, calculated using the object's signed distance field~($\Phi$) and the fingertips' positions $\boldsymbol{p}_{1:m}$ obtained from forward kinematics. Specifically, for any 3D point $\boldsymbol{x}$, we locate the nearest points $\bar{\boldsymbol{p}}$ on the object's surface and the outward normal $\bar{\boldsymbol{n}}$. $\Phi(\boldsymbol{p}) = \bar{\boldsymbol{n}} \cdot (\boldsymbol{p} - \bar{\boldsymbol{p}})$, with $\frac{\partial \Phi(\boldsymbol{x})}{\partial \boldsymbol{x}} = \bar{\boldsymbol{n}}$. 

%Given any 3d point $\boldsymbol{x}$, $\Phi(\boldsymbol{x})$ finds the nearest point $\bar{\boldsymbol{p}}$ on the object's surface from $\boldsymbol{x}$ and its outwards normal $\bar{\boldsymbol{n}}$, providing $\Phi(\boldsymbol{p}) = \bar{\boldsymbol{n}} \cdot (\boldsymbol{p} - \bar{\boldsymbol{p}})$.
%As for the second term, we introduce the object's signed distance field $\Phi$. Given any 3d point $\boldsymbol{x}$, $\Phi(\boldsymbol{x})$ finds its nearest point $\bar{\boldsymbol{x}}$ on the object's surface and the outwards normal $\bar{\boldsymbol{n}}$ at $\bar{\boldsymbol{x}}$, providing $\Phi(\boldsymbol{x}) = \bar{\boldsymbol{n}} \cdot (\boldsymbol{x} - \bar{\boldsymbol{x}})$ and $\frac{\partial \Phi(\boldsymbol{x})}{\partial \boldsymbol{x}} = \bar{\boldsymbol{n}}$. While $\boldsymbol{p}_{1:m}(\boldsymbol{q})$ denote the center positions of $m$ fingertips obtained from forward kinematics, the second term is then the sum of distances between fingertips and the object's surface.

We also add a set of constraints to make the resulting hand pose practical. Eqn.~\ref{eqn:c_limit} describes the robot joint limits. For simplicity, we use a set of $n$ spheres with different radius $\epsilon_{1:n}$ to represent the collision shape of the robot hand. The spheres are attached to the fingers and palm, so the centers' positions $\boldsymbol{c}_{1:n}(\boldsymbol{q})$ can be obtained from forward kinematics. Eqn.~\ref{eqn:c_penetration} guarantees no penetration between the hand and object and Eqn.~\ref{eqn:c_self_penetration} guarantees no self-penetration.

We solve this nonlinear optimization problem iteratively by performing Taylor expansion to Eqn.~\ref{eqn:objective}-\ref{eqn:c_self_penetration} in the neighborhood of $\boldsymbol{q}$, $\boldsymbol{f}_{1:m}^*$ and obtaining a convex optimization problem (second-order cone programming) which could be solved quickly by CVXPY~\cite{diamond2016cvxpy} as follows: \vspace{-10pt}

{\footnotesize
\begin{align}
    \min_{\delta \boldsymbol{q}, \delta \boldsymbol{f}_{1:m}} \; &  \left\Vert \boldsymbol{r}^*(\boldsymbol{w}, \boldsymbol{q}) + \sum_{i=1}^m \frac{\partial G_i(\boldsymbol{q})}{\partial \boldsymbol{q}} \boldsymbol{f}_i^* \delta \boldsymbol{q} 
    + G_i(\boldsymbol{q}) \delta \boldsymbol{f}_i \right\Vert ^2 \\
    &+ \sum_{i=1}^m \Big\vert \Phi(\boldsymbol{p}_i) + \frac{\partial \Phi(\boldsymbol{p}_i)}{\partial \boldsymbol{p}_i} \frac{\partial \boldsymbol{p}_i(\boldsymbol{q})}{\partial \boldsymbol{q}} \delta \boldsymbol{q} \Big\vert ^2, \\ 
    \textrm{s.t. }& \boldsymbol{f}_{1:m}^* + \delta \boldsymbol{f}_{1:m} \in FC(\mu), \\
    &\boldsymbol{q} + \delta \boldsymbol{q} \in [\boldsymbol{q}_{min}, \boldsymbol{q}_{max}], \\
    &\Phi(\boldsymbol{c}_{1:n}) + \frac{\partial \Phi(\boldsymbol{c}_{1:n})}{\partial \boldsymbol{c}_{1:n}} \frac{\partial \boldsymbol{c}_{1:n} (\boldsymbol{q})}{\partial \boldsymbol{q}} \delta \boldsymbol{q} \ge \epsilon_{1:n}, \\
    &\Big\Vert \boldsymbol{c}_i - \boldsymbol{c}_j + \big(\frac{\partial \boldsymbol{c}_i(\boldsymbol{q})}{\partial \boldsymbol{q}} - \frac{\partial \boldsymbol{c}_j(\boldsymbol{q})}{\partial \boldsymbol{q}} \big) \delta \boldsymbol{q} \Big\Vert \ge \epsilon_i + \epsilon_j, \\
    & \big\vert \delta \boldsymbol{q} \big\vert \le s_q ,\; 
    \big\vert \delta \boldsymbol{f_{1:m}} \big\vert \le s_f. \label{eqn.local}
\end{align}
}

\vspace{-3pt} In the above equations, the Jacobians $\frac{\partial \boldsymbol{p}(\boldsymbol{q})}{\partial \boldsymbol{q}}$ and $\frac{\partial \boldsymbol{c}(\boldsymbol{q})}{\partial \boldsymbol{q}}$ are provided by the simulation Jade~\cite{gang2023jade}. 
The exact formula of grasp mapping matrices and the corresponding gradients are derived in supplementary. Besides the constraints derived from the original problem Eqn.~\ref{eqn:objective}-\ref{eqn:c_self_penetration}, we also add Eqn.~\ref{eqn.local} restricting $\delta \boldsymbol{q}$ and $\delta \boldsymbol{f_{1:m}}$ in the neighborhood of zero, so that Taylor expansion is reasonable. We then update the joint values of the robot hand by $\boldsymbol{q} \leftarrow \boldsymbol{q} + \delta \boldsymbol{q}^*$. The convergence conditions are $\Vert \boldsymbol{r}^*(\boldsymbol{w}, \boldsymbol{q}) \Vert < 10^{-6}$ and $\vert \Phi \big(\boldsymbol{p}_{1:m}(\boldsymbol{q}) \big) \vert < 0.005$. We set the maximum number of iterations as 100. The optimization is successful if it reaches convergence within 100 iterations, otherwise, it fails and terminates after 100 iterations. The successful output of this iterative convex optimization is the joint values of the robot hand and associated torques to generate the wrench closest to the target wrench.
%Our optimization also supports region selection as described in Sec.~\ref{sec:contactSelection}, when only part of the object's surface is available for fingertips to touch because of human preference or obstacles in the environment. The signed distance field $\Phi$ in Eqn.~\ref{eqn:objective} should be replaced by $\Phi^R$, which calculates the minimum distance to the available region $\Phi^R(\boldsymbol{x})$ from any 3d point $\boldsymbol{x}$, guaranteeing the fingertips only touch the available region when the optimization converges. Since the available region is no longer a watertight manifold, $\Phi^R$ is not a signed distance field. Let $\bar{\boldsymbol{x}}^R$ be the nearest point on the available region from $\boldsymbol{x}$, $\Phi^R(\boldsymbol{x}) = \vert \boldsymbol{x} - \bar{\boldsymbol{x}}^R \vert$ and $\frac{\partial \Phi^R(\boldsymbol{x})}{\partial \boldsymbol{x}} = \frac{\boldsymbol{x} - \bar{\boldsymbol{x}}^R}{\vert \boldsymbol{x} - \bar{\boldsymbol{x}}^R \vert}$.

%\lin{ToDo for Yang Gang: Relationship to Force-closure Grasp}
We can provide force-closure grasp optimization by setting $\boldsymbol{w} = [0, 0, 0, 0, 0, 0]^T$, because when the supporting wrench set of the resulting grasp contains the neighborhood of origin point in wrench space, it can also expand to the whole wrench space by scaling the contact forces~\cite{boyd2007fast}.

The DoF (degree of freedom) for articulated rigid bodies is less than six (wrench space). So we extend the definition of wrench residual to joint space, $\boldsymbol{r}^J(P^J, \boldsymbol{w}, \boldsymbol{q}, \boldsymbol{f}_{1:m}) = \sum_{i=1}^m P^J \big( G_i(\boldsymbol{q}) \boldsymbol{f}_i \big) - \boldsymbol{\tau}$, where $\boldsymbol{\tau}$ is the target joint torque and $P^J$ is the projection matrix. If we apply a wrench $\boldsymbol{w}$ to the articulated object, the resulting torque at the joint will be $P^J \boldsymbol{w}$. Take the revolution joint as an example, where $\boldsymbol{a}$ is the axis direction and $\boldsymbol{o}$ is any point on the axis. The corresponding projection matrix is $P^J = \big[ (\boldsymbol{o} \times \boldsymbol{a})^T, \boldsymbol{a}^T \big]$. After replacing the definition of wrench residual, the other parts of optimization are the same as free objects.

\section{Dataset}
We create a large-scale comprehensive annotated dataset that includes articulated/rigid objects and deformable objects in 1D/2D/3D forms. Refer to our ~\href{https://manifoundationmodel.github.io/}{website} for more details.
%For articulated/rigid objects, over 100K objects from multiple sources were normalized, scaled, and convex decomposed to annotate contact regions and forces generated with eight kinds of common robot hands. For 2D deformable objects, more than 2K models from \textit{ClothesNet}~\cite{zhou2023clothesnet} were simulated in \textit{DiffCloth}~\cite{li2022diffcloth} to record their motion and physical properties. For 3D and 1D deformable objects, both 1K models were simulated in SoftMac~\cite{liu2023softmac} to collect their deformation under applied forces. See our website for more details.tt

\paragraph{Training Set} For fair and convenient network comparison, we select a small $\sim$3K training dataset, consisting of $\sim$1K rigid, $\sim$1K clothes, and $\sim$1K 3D deformation object examples. Each example contains the input of our network: oriented object point cloud, target point cloud, target motion, object physical properties, manipulation region mask, manipulator point cloud, and the ground truth contact point heatmap. The rigid dataset part contains data generated with PandaGripper, KinovaHand, and LeapHand. We also have a $\sim$200K training dataset to evaluate whether our model can scale up with more data.  

\paragraph{Testing Set} To systematically evaluate the performance, we design $2\times3=6$ test sets. The ``3'' here stands for three types of objects (rigid, cloth, 3D deformable). The ``2'' here is for two test groups: Seen Object \& Unseen Motion and Unseen Object \& Unseen Motion. Specifically, the Seen Object \& Unseen Motion test group is designed to assess whether our model can learn to manipulate an object to reach an unseen arbitrary goal after seeing the object in the training set. The Unseen Object \& Unseen Motion test group is designed to evaluate our model's generalization to novel objects. Each test set contains $\sim$200 examples.

\section{Experiments}
In this section, we conduct experiments to answer the following questions:
\begin{itemize}
    \item[] \hspace{-2em}Q1: What is the overall performance of our model?
    \item[] \hspace{-2em}Q2: How does training on different object categories affect our performance?
    \item[] \hspace{-2em}Q3: How do physical properties affect the model output?
    \item[] \hspace{-2em}Q4: Can our CVAE-based network generate diverse proposals with the same input?
    \item[] \hspace{-2em}Q5: Can manipulation region mask constrain the network contact heatmaps and the optimization result?
    \item[] \hspace{-2em}Q6: How effective is our grasp optimization? 
    \item[] \hspace{-2em}Q7: Does combining network and optimization improve task-oriented grasp generation? (See Supp.)
    \item[] \hspace{-2em}Q8: Can our model generalize to novel robot manipulators?  (See Supp.)
    \item[] \hspace{-2em}Q9: How does our proposed framework perform on real robot experiments with real point clouds? (See Supp.)
\end{itemize}
To report quantitatively results, we evaluate our methods by success rate all in simulations. Success for a test example is defined for each material as follows: 

\textbf{For (articulated) rigid objects}, 
    % \gang{TODO}. With predicted points in hand, we first use ICP (Iterative Closest Point) algorithm to obtain every initial hand pose for different finger correspondence. Then we can generate a heatmap on the object point cloud from the fingertips position for each initial hand pose. Based on the IoU (Intersection over Union) between the generated heatmap and the predicted contact point heatmap, we can find the best matching initial pose. Sending the chosen initial pose to our Rigid-body Grasp Refinement, an wrench-based optimization can be finally implemented. 
    Success is (1) finding a hand pose and finger joint values to grasp the object without penetration; (2) moving the object with the optimized contact force in simulation assuming precise mass and inertia, reaching a 6D pose whose average point-wise distance to the target point cloud is no more than 1mm. %In the simulation, we assume precise mass and inertia. %, but in practice, we care more about the direction of wrench and leave low-level force/torque control to move towards target shape.
    % selected contact points can produce contact forces that can span the target wrench with a mean square error $<10^{-6}$.

\textbf{For deformable objects}: We apply the predicted contacts on the objects in simulation. We use DiffCloth~\cite{li2022diffcloth} for 2D deformable objects and SoftMAC~\cite{liu2023softmac} for 1D/3D deformable objects. We apply contact forces to the object and obtain the rollout point cloud in simulators. The success is determined by the distance between the rollout point cloud and the target point cloud. Detailed thresholds and parameters are in Supp.
%\lin{Need to rewrite}
    
%Initially, the distance differences between the two sets are calculated, and the top 50 points with the largest distance differences are selected for further evaluation. The test result is deemed a failure if the distance of any of these points exceeds 0.5. Furthermore, the test result is also considered a failure if more than 50\% of these points have a distance greater than 0.3. 
%For 3D deformable objects, we also select 100 points with the largest distance difference between the initial and target point clouds. Then we choose several contact points and forces according to model prediction, applying the forces to the chosen contact points to get a predicted point cloud. We compare the predicted point cloud with the ground truth target point cloud. A prediction is denoted as success if the average L2 distance between the predicted point cloud and the target point cloud is less than 0.5cm, and the average distance of the selected 100 points is less than 5cm.

\subsection{Neural Network Evaluation}

\subsubsection{Overall Performance \& Data Fusion}~\label{sec:q1}

\begin{table}[t]
\resizebox{0.48\textwidth}{!}{
\centering
\begin{tabular}{lcccccc}
\toprule
\multirow{2}{*}{Methods SR} & \multicolumn{3}{c}{Seen Object \& Unseen Motion} & \multicolumn{3}{c}{Unseen Object \& Unseen Motion} \\ \cline{2-7}
& Rigid & 3D Deform & 2D Deform & Rigid & 3D Deform & 2D Deform \\ \midrule
Baseline            & 23.5\%         & 2.0\%              & 0.6\%              & 22.5\%         & 2.5\%              & 3.0\%              \\ 
Ours(SD)            & \textbf{92.5\%}         & 88.5\%             & 86.23\%            & 97.0\%         & 82.0\%             & 84.0\%             \\ 
Ours(3K)            & 91.0\%         & 85.0\%             & \textbf{89.22\% }           & 96.0\%         & 82.5\%             & 82.5\%             \\ 
Ours                & 91.5\%         & \textbf{91.0\%}             & 83.23\%            & \textbf{97.5\%}       & \textbf{85.5\%}            & \textbf{84.5\%}             \\ \bottomrule 
\end{tabular}
}
\caption{Success rates of the network over different experiment settings.\vspace{-25pt}}
\label{tab:neteval}
\end{table}

\begin{figure}[t]
\centering
\includegraphics[width=0.48\textwidth]{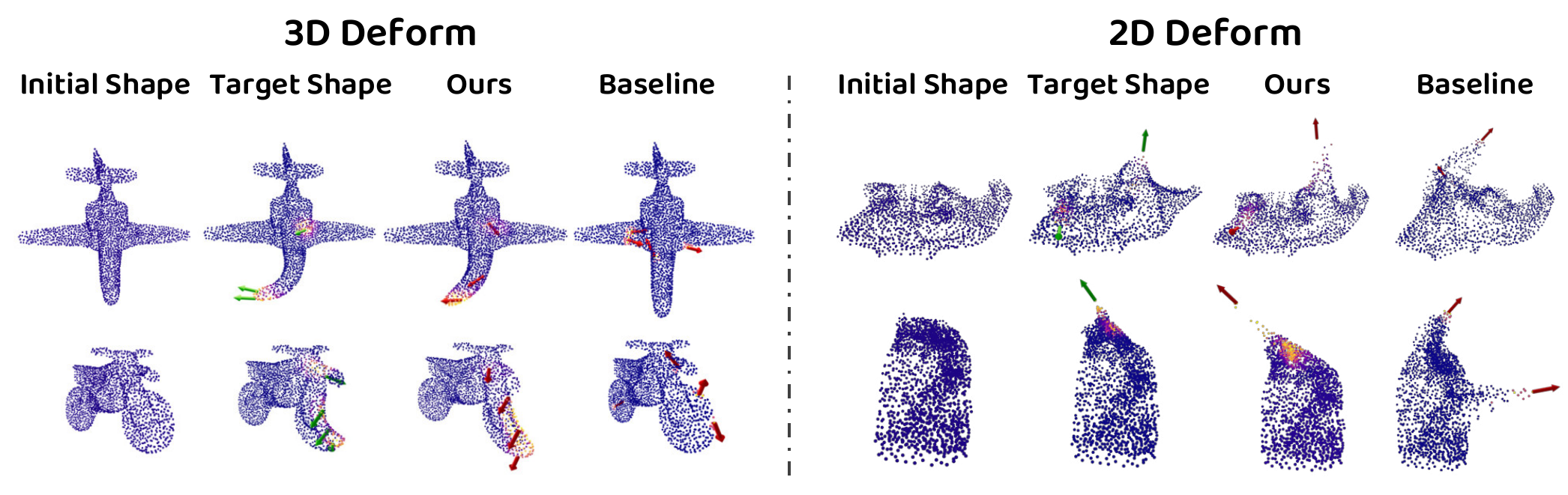}
\vspace{-20pt}
\caption{Visualizations of evaluations on deformable objects are shown. The point cloud colors represent the contact heatmap, with purple indicating low values and yellow indicating high values. Arrows denote the force or motion direction. This representation is consistent across all figures.}
\vspace{-20pt}
\label{fig:deform_eval}
\end{figure}

To answer Q1 and Q2, we train our model described in Sec.~\ref{sec:NN_arch} on the above 3K training set denoted as \textbf{Ours(3K)} and compare it with three approaches: \textbf{Baseline}, \textbf{Ours(SD)}, and \textbf{Ours}. \textbf{Baseline} randomly selects the same number of contact points as \textbf{Ours}, applying forces or motions in random directions for deformable objects. For articulated/rigid object, \textbf{Baseline} sets the robot hand/fingers at the maximum open status, then moves it toward the object's center, closing the fingers upon contact with the object's surface. \textbf{Baseline} calculate the forces of each fingerstip using Eqn.~\ref{eqn:wrenches}. 
%applies  random grasp, drag, or press on several random points on the input point cloud for 2D and 3D deformable objects.\lin{explain the baseline for rigid body} 
We chose such a baseline because we are the first to train a unified model for contact synthesis on objects of various types. \textbf{Ours(SD)} stands for training our model only on Single Dataset ($\sim$1K rigid, $\sim$1K clothes, $\sim$1K 3D deformation respectively instead of together). \textbf{Ours} is for our network trained on a $\sim$200K dataset training data. 

To answer Q1, as shown in Tab. \ref{tab:neteval}, \textbf{Ours} has an average success rate of over 90, 85, 80\% over rigid, 3D deformable, and 2D deformable objects. It indicates our pipeline can synthesize reasonable contacts for diverse objects with broad generalization. Meanwhile, the low success rate of \textbf{Baseline} and the visualization \ref{fig:deform_eval} illustrate the difficulty of each task. %Moreover, all of \textbf{Ours(SD)}, \textbf{Ours(3K)} and \textbf{Ours} do not have a significant performance loss on the test set of seen objects, on the contrary, some metrics are even higher. It indicates the strong generalization ability of our network design. Moreover, \textbf{Ours} have better or comparable performance to \textbf{Ours(3K)}, especially on 3D deformation objects, showing that our model has the potential to scale up with a larger dataset.\lin{Still has issues} 

To address Question 2, we conducted a comparison between \textbf{Ours(3K)} and \textbf{Ours(SD)}. The \textbf{Ours(3K)} have comparable or slightly better performance. This reflects that our model is capable of effectively integrating various object types without performance loss, compared to training on a single object type. %We hypothesize that the three kinds of objects can be successfully trained together for the following reasons. First, the formulation of our foundation manipulation problem makes it easier for the network to learn and generalize. Second, our network has physical attributes as inputs to the network, which help the network to discriminate between the three objects. Third, the target motion of the objects somehow contains information that can be used to distinguish the properties of the objects. 
\subsubsection{Physical Properties}
\begin{figure}[thb!]
\centering
\includegraphics[width=0.48\textwidth]{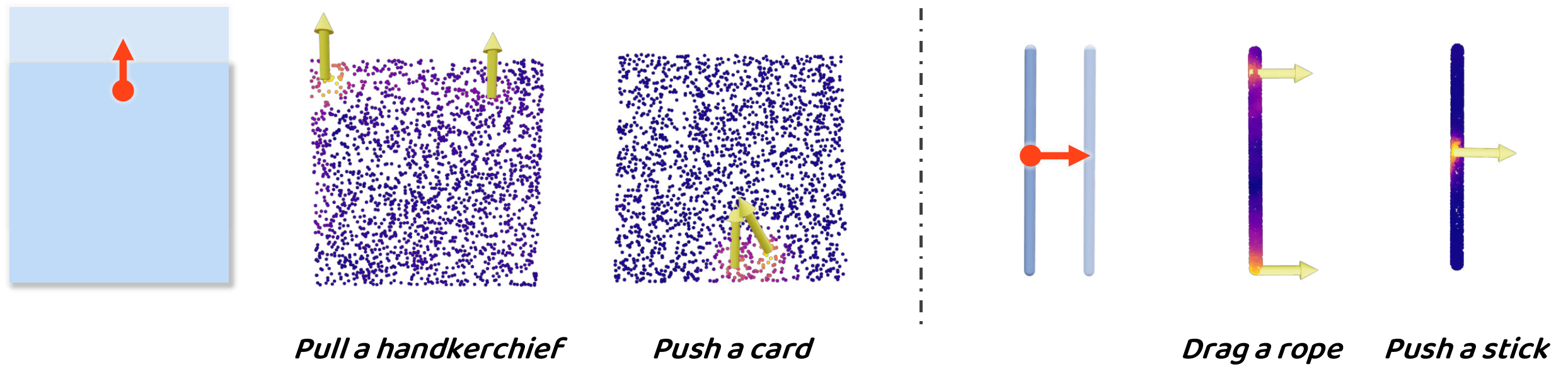}
\vspace{-20pt}
\caption{Visualizations of how physical properties affect network outputs. Blue squares and cylinders are objects. Overlapping images with transparency represent the moved object. The orange arrow represents the motion.}
\vspace{-20pt}
\label{fig:mat}
\end{figure}

To understand the impact of physical properties (in Q3), we conducted two experiments shown in Fig.~\ref{fig:mat}, evaluating how variations in physical properties influence our model’s behavior. We used two shapes: a thin square and a long cylinder. The square represented a handkerchief (deformable) or a card (rigid). The cylinder represented a rope (deformable) or a stick (rigid). We used identical point clouds and motions, altering only the physical properties to observe the outcomes. Fig.~\ref{fig:mat} shows that for the thin square, our model pulls the forward corners for the handkerchief, reflecting its pliability, and pushes from the rear for the rigid card. For the long cylinder, our model drags both ends of the rope and pushes the middle of the stick. These results demonstrate that our model adjusts its outputs based on the physical properties, indicating it learns to reason about object dynamics. 
 
\subsubsection{Multimodal Syntheses}
\begin{figure}[thb!]
\centering
\vspace{-10pt}
\includegraphics[width=0.48\textwidth]{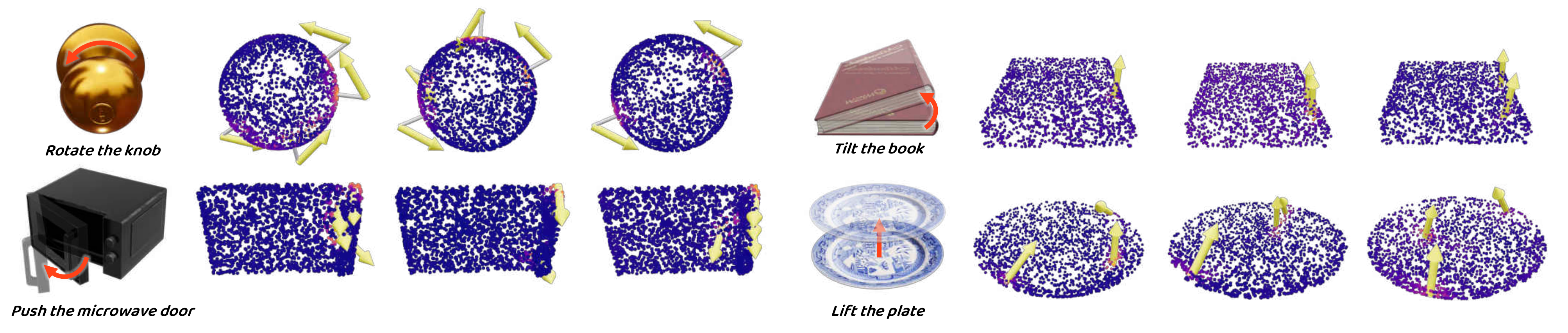}
\vspace{-20pt}
\caption{Visualization of multimodal contact syntheses generated by our CVAE-based model.}
\label{fig:cvae}
\vspace{-10pt}
\end{figure}

To answer Q4—whether our model can generate diverse solutions from the same input—we tested four novel objects not in our training dataset. For each object, we selected a common motion, such as rotating a knob or tilting a book. Using these inputs, we sampled different $z\in \mathbb{R}^{64}$ values from $\mathcal{N}(0,I)$ and visualized the outputs. As shown in Fig. \ref{fig:cvae}, our CVAE module allows our model to generate diverse, physically plausible contact outputs. For example, to rotate a knob, the model provides solutions like holding the upper or the lower side.

\subsubsection{Manipulation Region Selection} 
\begin{figure}[thb!]
\centering
\vspace{-10pt}
\includegraphics[width=0.48\textwidth]{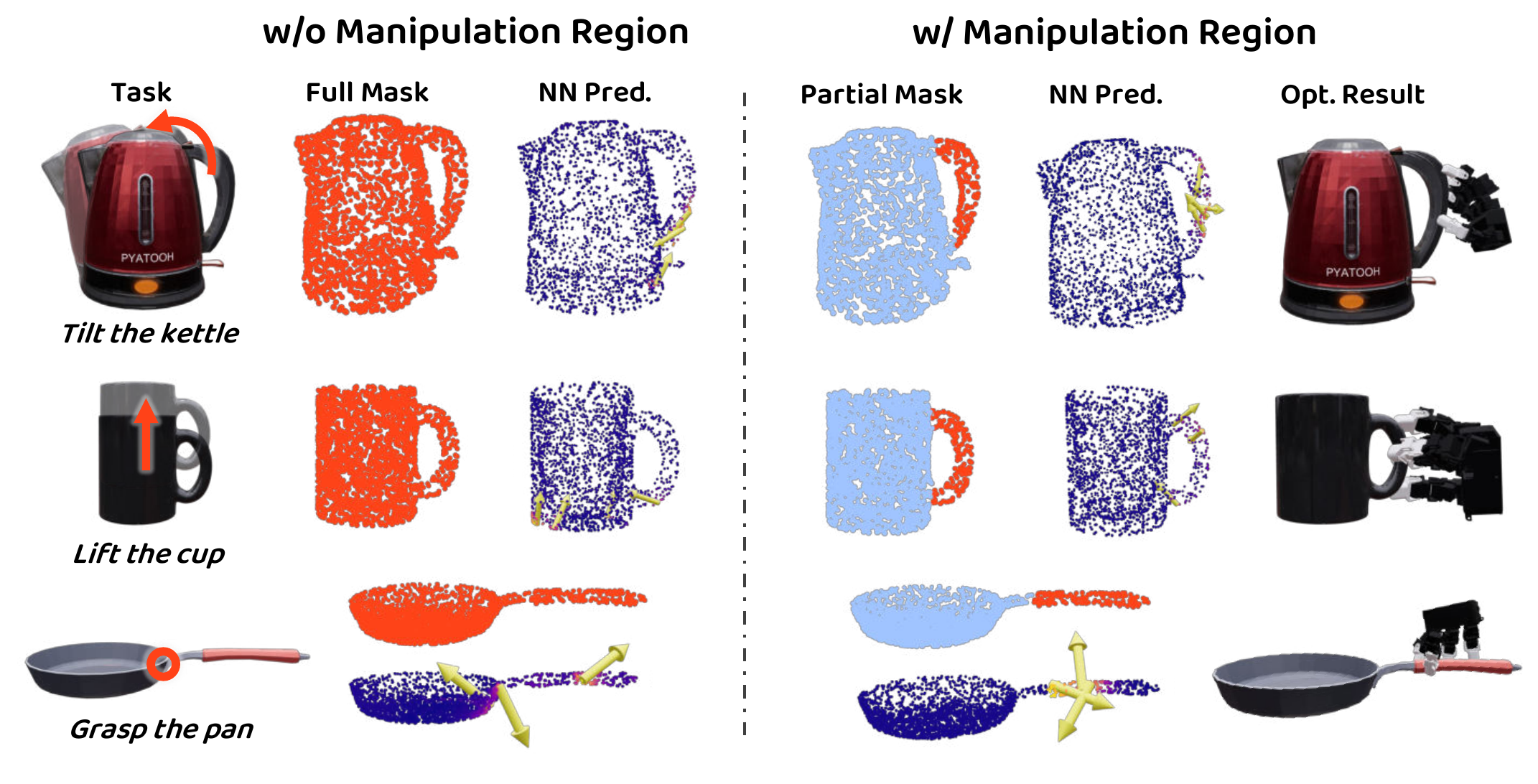}
\vspace{-20pt}
\caption{Visualizations of how manipulation region mask input affects our model prediction and the optimization result.}
\label{fig:region}
\vspace{-10pt}
\end{figure}

To explore whether a manipulation region mask can guide the network's contact point predictions (Q5), we tested our model with three new everyday objects and tasks not included in our training set. These objects have specific manipulation areas, like handles. We assessed the model's performance with and without a manually selected manipulation region mask, and the results are shown in Fig. \ref{fig:region}. The mask guides the model to favor these regions. Without a mask, it may suggest contact points suitable for the intended motion but impractical for manipulation. For instance, lifting a cup from its bottom on a table is unfeasible. Applying a mask to the handle helps the model propose viable contact points, preventing potential collisions. The effectiveness of this approach is illustrated by the robot hand poses generated by our model, shown on the right side of Fig. \ref{fig:region}. Although generating the ideal manipulation mask isn't our focus, our model includes an API for users to define their manipulation regions.

\subsection{Evaluating Robot Hand Pose Optimization}
We evaluate the effectiveness of our proposed robotic hand pose refinement for task-oriented and force-closure grasps. Task-oriented grasp requires the grasp pose to produce a specific target wrench.  As derived in Sec.~\ref{sec:opt}, we use the target wrench as the manipulation task description for rigid objects. Our evaluation includes both task-oriented and force-closure grasp scenarios over 1000 objects in our dataset, generating 20 distinct grasps per object, totaling 20k task-oriented and 20k force-closure grasps. In task-oriented evaluations, the target wrench for each grasp is randomly sampled. A grasp is considered successful if it meets three criteria: the residual wrench is less than $10^{-6}$, the distance between the fingertip and the object is less than 5mm, and there are no intersections between fingers or between fingers and the object.

For a fair comparison, we initialize each method identically. We retrieve a ground truth 6D palm pose from our dataset that can exert the target wrench or achieve force-closure. This 6D palm pose is placed 30cm away from the object, and then a random 60-degree rotation and 5cm translation are added, with all finger joints set at rest positions. Under this initialization, we run our robotic hand pose refinement and compare it with a direct grasp method. The direct grasp method moves the robot hand directly towards the object's center, closes the fingers upon contact, and calculates the forces for each fingertip using Eqn.~\ref{eqn:wrenches}.

%our wrench-based hand pose refinement module, and compare it with a baseline direct-grasp method described. To establish the baseline, we simplify our approach by excluding the wrench-matching pattern and merely minimizing the distance between the fingertips and the object's surface, mimicking a straightforward gripping action. 

%We evaluate the effectiveness of each method by 1) starting from an initial hand pose by each method, 2) applying the refinement process and 3) assessing the feasibility of the resulting hand pose as in 
Our analysis includes three types of robotic hands: Barrett~\cite{barrett}, LeapHand~\cite{shaw2023leaphand}, and ShadowHand~\cite{shadow}. As shown in Table \ref{tab:graspeval}, our refinement module achieves significantly higher success rates than the baseline for both scenarios and all tested hands.

%The results indicates the our proposed robotic hand pose refinement can generate grasp poses for general grasping/manipulation tasks. 
\begin{table}[t]
\centering
\resizebox{0.48\textwidth}{!}{%
\begin{tabular}{cclclcl}
\toprule
\multirow{2}{*}{Method} & \multicolumn{2}{c}{Barrett}          & \multicolumn{2}{c}{LeapHand}         & \multicolumn{2}{c}{Shadow}           \\ \cmidrule{2-7} 
                        & \multicolumn{3}{r|}{Task-oriented SR}                              & \multicolumn{3}{l}{Force-closure SR}          \\ \midrule
Baseline                    & \multicolumn{1}{c|}{T 16.5\%} & F 13.2\% & \multicolumn{1}{c|}{T 22.1\%} & F 23.7\% & \multicolumn{1}{c|}{T 10.4\%} &F 18.9\% \\ 
\midrule
Ours                & \multicolumn{1}{c|}{T 95.2\%} &F 96.3\% & \multicolumn{1}{c|}{T 91.9\%} & F 95.1\% & \multicolumn{1}{c|}{T 77.7\%} &F 81.9\% \\ \bottomrule
\end{tabular}}
\caption{Hand pose refinement and direct grasp comparisons. T and F indicate task-oriented grasp and force-closure grasp.}
\label{tab:graspeval}
\vspace{-30pt}
\end{table}

\subsection{Real World Experiments}
%\begin{figure}[H]
%\centering
%\includegraphics[width=0.48\textwidth]{img/realworldexp_c.pdf}
%\caption{Realworld experiment settings.}
%\label{fig:realworld}
%\end{figure}
In this section, we evaluate our system's performance in manipulating various rigid objects, clothes, and 3D deformable objects in real-world settings, as shown in Fig. \ref{fig:figure1}. Detailed descriptions are provided in the Supp. \textbf{Rope Rearrangement} We designed an experiment for a Kinova MOVO robot arm to rearrange a randomly reset rope into the letters "NUS." We performed the task 10 times for each letter, achieving a 90\% success rate, indicating our model's accuracy in generating grasp points for deformable objects like ropes. \textbf{Breakfast Preparation}
We designed a complex breakfast preparation experiment using a single Flexiv robot arm with a LeapHand manipulator. The tasks included opening the fridge door (articulated object), taking out the milk box (rigid object), placing the milk box on the table, picking up a piece of bread, putting it into the toaster, opening the toaster, and placing the cooked bread on a plate. The bread is treated as a 3D deformable object. \textbf{Cloth Folding} We designed a cloth folding experiment for a Kinova MOVO robot arm to fold a T-shirt. The robot grasps the point with the highest heatmap value and moves to the predicted target point.
%This action will be executed repeatedly until all points in the rope point cloud have a max $L2$ error $<0.05m$ and a mean $L2$ error $<0.03m$ w.r.t. its target point in this stage. 

\section{Conclusion and Limitations}
\label{sec:conclusion}
We introduce a framework for developing a manipulation foundation model for contact synthesis. This model takes inputs like point clouds, physical properties, target motions, and region masks to output contact points and forces or motions. Tested on various objects, including articulated, rigid, and deformable items, our model achieves average success rates of around 90\% in both simulations and real-world settings.

Our model can be improved in the following aspects. First, our framework focuses on quasi-static tasks by separating the trajectory into discrete motions at each time step. Future work could extend the framework to high-dynamic tasks by considering multi-step motions or the entire trajectory at each timestep. Second, our model focuses on single-hand finger-point contact. Future work could extend the framework to multi-hand surface contact situations. Thirdly, our current model has about 5 million parameters. In the future, we plan to scale up by increasing the amount of data and the number of network parameters.

%{\scriptsize

{\small
\bibliographystyle{IEEEtran}
\bibliography{references}
}
%{\scriptsize
%\bibliographystyle{IEEEtran}
%\bibliography{references}
%}

\clearpage
\appendix

\section{Technical Approach}

\subsection{High-level Task and Motion Planning via LLM}
In our work, we develop an LLM/LVM-based object motion planner that can generate collision-free dense trajectories that follow high-level language instructions $\mathcal{I}$ from users that are long-horizon and require contextual understanding of the embodied world. 
In our work, the planner consists of the following components: 
1) an LLM/LVM that is prompted with the description of the task to generate sparse 6D arrays that might contain infeasible waypoints or collision, 2) An open-vocabulary object detector to obtain spatial-geometrical information of the relevant objects, or even portions of the object (e.g. ``the body of the microwave" or ``the handle of the cup"), and 3) a sampling-based motion planner to generate collision-free dense trajectory.
To obtain a sparse trajectory that roughly describes the pose change of manipulated objects from the initial states to the goal states, the planner first transforms the $\mathcal{I}$ (e.g. ``Put the cup in the microwave" and ``Pull the rope into the shape of N, U and S") into several decompositions $\mathcal{I} \rightarrow (i_{1}, i_{2}, ..., i_{n})$.
To obtain precise geometrical information about each object, We employ the multimodal detector~\cite{ren2024grounded} to generate 2D bounding boxes around each object, using object names as prompts, these 2D coordinates are then projected onto 3D space to gather the spatial-geometry information corresponding to the robot perspective coordinates.
With the augmentation geometric information of objects, the LLM then generates a trajectory $\tau_{i}$ of the manipulated objects for each manipulation phase described by the sub-instruction $i_{t}$ in a chain-of-thought~\cite{codeaspolicies2022, yu2023multireact} style, where each waypoint consists of 6-DoF object pose.
However, trajectories proposed by the LLM are not entirely reliable, as the text description and image of the environment does not contain all the comprehensive physics information of the embodied environment, and as the environment is not fully observable to the LLM/VLM, so that the waypoints may contain infeasible states or collisions with the surrounding environment. 
To avoid giving the foundation model infeasible waypoints, we adopt a sampling-based motion planning method to interpolate the sparse trajectory $\tau_{i}$ to a collision-free and feasible dense trajectory
$OPT(\tau_{i})$.

\subsection{Middle-level Motion Planning via Simulation}
To enable flexible collision checking while motion planning, we represent all the objects with their point clouds and load them in the Jade simulator~\cite{gang2023jade}.
We adopt fcl~\cite{6225337} for real-time collision checking for objects with the surroundings in the simulation environment.
We use the following RRT-connect motion planning algorithm ~\ref{rrt} to find a collision-free dense trajectory. 
\begin{algorithm}
\caption{Generate Collision-Free Dense Trajectory using RRT-Connect}
\label{rrt}
\begin{algorithmic}
\REQUIRE Sparse trajectory $\tau_{i}$ from LLM/LVM, initial state $s_{\text{init}}$, goal state $s_{\text{goal}}$, obstacle set $\mathcal{O}$
\ENSURE Dense trajectory $OPT(\tau_{i})$ that is collision-free
\STATE Initialize $T_{\text{start}}$ with root $s_{\text{init}}$
\STATE Initialize $T_{\text{goal}}$ with root $s_{\text{goal}}$
\WHILE{$T_{\text{start}}$ and $T_{\text{goal}}$ not connected}
    \STATE $s_{\text{rand}} \gets$ SampleRandomState()
    \STATE $s_{\text{near}} \gets$ NearestNeighbor($T_{\text{start}}$, $s_{\text{rand}}$)
    \STATE $s_{\text{new}} \gets$ Steer($s_{\text{near}}$, $s_{\text{rand}}$)
    \IF{not InCollision($s_{\text{new}}$, $\mathcal{O}$)}
        \STATE Add $s_{\text{new}}$ to $T_{\text{start}}$
        \IF{CanConnect($s_{\text{new}}$, $T_{\text{goal}}$)}
            \STATE Connect $T_{\text{start}}$ and $T_{\text{goal}}$ through $s_{\text{new}}$
        \ENDIF
    \ENDIF
    \STATE Swap($T_{\text{start}}$, $T_{\text{goal}}$)
\ENDWHILE
\STATE $OPT(\tau_{i}) \gets$ PathBetweenTrees($T_{\text{start}}$, $T_{\text{goal}}$)
\STATE $OPT(\tau_{i}) \gets$ OptimizePath($OPT(\tau_{i})$)
\STATE \Return $OPT(\tau_{i})$
\end{algorithmic}
\end{algorithm}
\section{Dataset}

\begin{figure*}[t]
\centering
\includegraphics[width=1.0\textwidth]{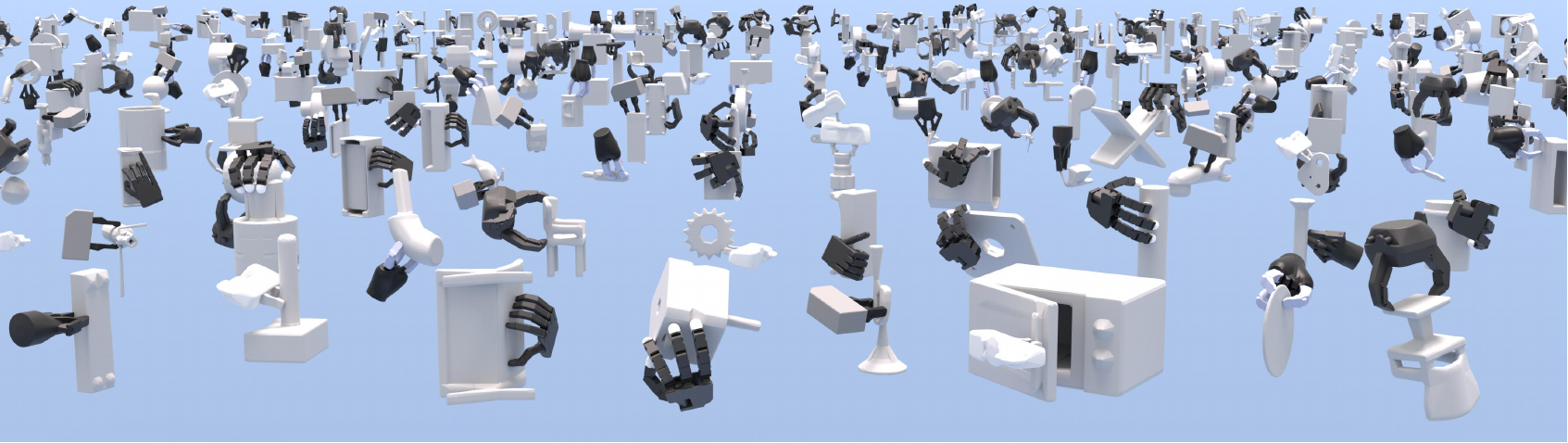}\vspace{4pt}
\includegraphics[width=1.0\textwidth]{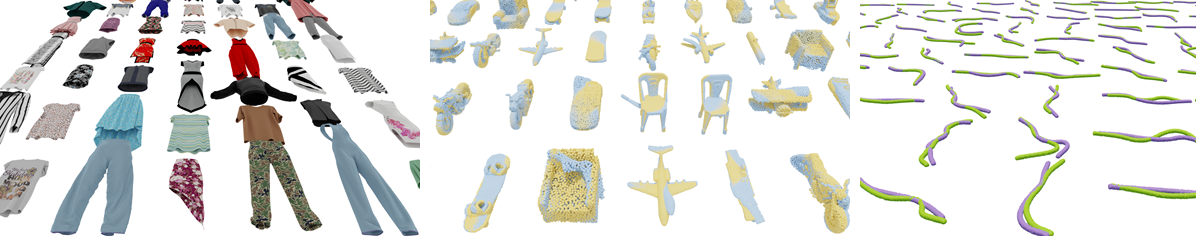}
\vspace{-10pt}
\label{fig:dataset}
\caption{Dataset visualization. Our large-scale comprehensive annotated dataset includes articulated/rigid and deformable objects in 1D/2D/3D forms.}
\end{figure*}

We create a large-scale comprehensive annotated dataset that includes articulated/rigid objects and deformable objects in 1D/2D/3D forms, as shown in \ref{fig:dataset}. 

\subsection{Articulated/Rigid Bodies}
\label{appendix:rigid}

\textbf{Object Pre-processing}
We collect 100K+ object models from 1K+ categories in Objaverse\cite{objaverse}, ShapeNet\cite{chang2015shapenet}, ABC\cite{koch2019abc}, Thingi10K\cite{Thingi10K}, and GAPartNet\cite{geng2023gapartnet}. From these datasets, we select 100K+ objects in 1K+ categories and normalize all models into a unit box and augment each object by randomly scaling them with 4 sizes between 0.05 and 0.4. Then we remesh them into manifolds\cite{huang2020manifoldplus}, abandon those with low volume and translate the obtained mesh so that its coordinate origin coincides with its center of mass. Finally, for simulation purposes, we create collision meshes for every object mesh through convex decomposition using CoACD\cite{wei2022approximate}. 

\textbf{Annotations} Our dataset contains millions of training examples. Each training example is composed of the input and the ground truth label. The input of each example is the object oriented point cloud $\mathcal{P}_o$, robot manipulator's point cloud $\mathcal{P}_h$ at its rest pose, task motion $\mathcal{M}$, manipulation region $\mathcal{R}$, and friction coefficient. The ground truth label contains the ground truth contact point heatmap and contact force heatmap. We propose a unified sampling-based method for large-scale contact synthesis. We adopted a hierarchical sampling approach to obtain suitable contact training examples. For a scaled object model, we sample points on its outer surface and obtain their corresponding normals. Since solving the inverse kinematics (IK) for multiple fingers with a floating base is relatively challenging and time consuming, we propose first sample palm poses facing the object, solve IK for each finger, and perform combinations on all finger solutions. We then sample contact forces within all the contact friction cones. At last, we calculate the sum of the wrenches exerted by all fingers to the object. Here we focus on point contact between the tips of robotic hands and the object surface. For the manipulation region defined as $\mathcal{R}=\left\{r_i\right\}_{i=1}^N$, there are 50\% probability that $r_i=1$ for all $i=1\dots N$, representing no constraints on the output contact heatmap.  
Conversely, the points where  $r_i=1$ encompass the ground truth contact points along with points in their vicinity.

\subsection{2D Deformable Object}
\label{appendix:deformable}

\textbf{Object Pre-processing}
We collect 2K+ clothes meshes from 4 categories and 19 sub-categories in ClothesNet \cite{zhou2023clothesnet}. For each clothes mesh, we normalize it into a unit box and simplify the mesh with quadric decimation \cite{garland1997surface} to under 3500 vertices. Objects with more than one connected component are filtered out. 

\textbf{Annotations}  Our dataset contains 100K+ training examples. Each training example has the same data format as rigid body. The physical properties of the cloth include its density, frictional coefficient, and elasticity coefficients. The manipulation region are all set to one.  Our 2D deformable dataset have two parts. One part is collected during clothes folding. The other is collected during random drags. Specifically, we first rotate the clothing so that the front side faces upwards, and let it free-falls in the DiffCloth \cite{li2022diffcloth} simulator until it lands flat on a plane, defining this as the initial object state. Then for the first part, we define a set of contact point based on key points detected by Skeleton Merger \cite{shi2021skeleton} and generate target motion trajectories for folding the cloth. 

\subsection{3D Deformable Object}
\label{appendix:3ddeformable}

\textbf{Object Pre-processing} For 3D Deformable objects, we collect 1K object models from 10 categories in ShapeNet~\cite{chang2015shapenet}. All the objects are normalized into a unit ball. Then, we use farthest point sampling (FPS) to sample 2048 points on the surface. We use SoftMAC~\cite{liu2023softmac}, a particle-based deformable object simulator, to simulate the motion of the objects when external forces are added.

\textbf{Annotations}
For each object, we collect 100 training examples. So, the total size of our 3D deformable object dataset is 100K. Each training example has the same data format as rigid body. The physics property includes friction coefficient, Young's modulus, and Poisson ratio. Our data annotation process is structured as follows: For an object composed of particles, we initially apply the K-means clustering method to divide all points into 100 clusters. Subsequently, we randomly select one group and then randomly choose a force to apply to all particles in this group. Clustering is adopted because the entire object is composed of 2048 particles. Applying a force to just one particle would have a minimal impact on the object's overall shape. Therefore, for a noticeable deformation in the entire object, we apply the same force to all particles within a selected cluster simultaneously.

% \begin{figure*}[t]
% \centering
% \includegraphics[width=0.38 \textwidth]{img/clothes.png}
% \hfill
% \includegraphics[width=0.38 \textwidth]{img/mpm3.png}
% \hfill
% \includegraphics[width=0.22 \textwidth]{img/rope.png}
% \caption{Overview of Dataset.}
% \end{figure*}

\section{Experiment}

\subsection{Simulation Experiments Metric for Deformable Objects}
We apply the predicted contacts on the objects in simulation. We use DiffCloth \cite{li2022diffcloth} for 2D deformable objects and SoftMAC \cite{liu2023softmac} for 1D/3D deformable objects. We can calculate the object point cloud after applying contact forces in simulators. The success rate is defined based on the distance between the results from the object point cloud and the target point cloud. 

Specifically, for 2D deformable objects, we evaluate using 50 points with the most motion. A prediction fails if the max L2 error exceeds 0.5m, or over 50\% of the selected points have an L2 error over 0.3m. For 3D deformable objects, we select 100 points. Success requires an average L2 error under 0.5cm and an average distance for selected points under 5cm. During evaluation, all point clouds are normalized into a cube with side length of 2m.

\begin{table}[t]
\centering
\begin{tabular}{ccccc}
\toprule
\multirow{2}{*}{Method}                                                           & \multicolumn{2}{c}{LeapHand}                                                                                & \multicolumn{2}{c}{Allegro}                                                                             \\ \cmidrule{2-5}
                                                          & \begin{tabular}[c]{@{}c@{}}SOUM \end{tabular} & \begin{tabular}[c]{@{}c@{}}UOUM \end{tabular} & \begin{tabular}[c]{@{}c@{}}SOUM \end{tabular} & \begin{tabular}[c]{@{}c@{}}UOUM \end{tabular} \\ \midrule
Baseline                                                   & 23.5\%                                            & 22.5\%                                            & 18.0\%                                            & 19.5\%                                            \\ \midrule
Ours                                                       & 91.5\%                                            & 97.5\%                                            & 94.0\%                                            & 94.0\%                                            \\ \midrule
\begin{tabular}[c]{@{}c@{}}Ours  w/o WBR\end{tabular}    & 24.0\%                                            & 24.0\%                                            & 17.0\%                                            & 16.5\%                                            \\ \midrule
\begin{tabular}[c]{@{}c@{}}Noisy GT w/ WBR\end{tabular} & 94.5\%                                            & 95.0\%                                            & -                                                  & -                                                  \\ \bottomrule
\end{tabular}
\caption{Rigid body grasp performance.}
\label{tab:q7q8}
\vspace{-10pt}
\end{table}

\subsection{Ablation Study on Robot Hand Pose Refinement}

For Q7 and Q8, we conduct the following experiments on the same rigid-body test set as the first experiment, with Seen Object \& Unseen Motion denoted as SOUM and Unseen Object \& Unseen Motion denoted as UOUM for simplicity. We report the success rate for LeapHand, as shown in Tab. \ref{tab:q7q8}, to answer Q7 and validate the performance with the couple of our network and wrench-based refinement. \textbf{Baseline} is the same as the first experiment, our wrench-based refinement is denoted as \textbf{WBR}. \textbf{Ours w/o WBR} replaces \textbf{WBR} with directly closing fingers as \textbf{Baseline} does, while \textbf{Noisy GT w/ WBR} stands for using ground truth hand poses with some Gaussian noise to enable our wrench-based refinement. As for the performance of our network, we can see that \textbf{WBR} significantly improves our network, based on the results of \textbf{Ours w/o WBR} and \textbf{Ours}. Furthermore, \textbf{Ours} gives a remarkably better performance than \textbf{Baseline}, and an at least not weaker performance than \textbf{Noisy GT} which can be regarded as a high-quality initialization.

\begin{figure}[H]
\vspace{-10pt}
\centering
\includegraphics[width=0.48\textwidth]{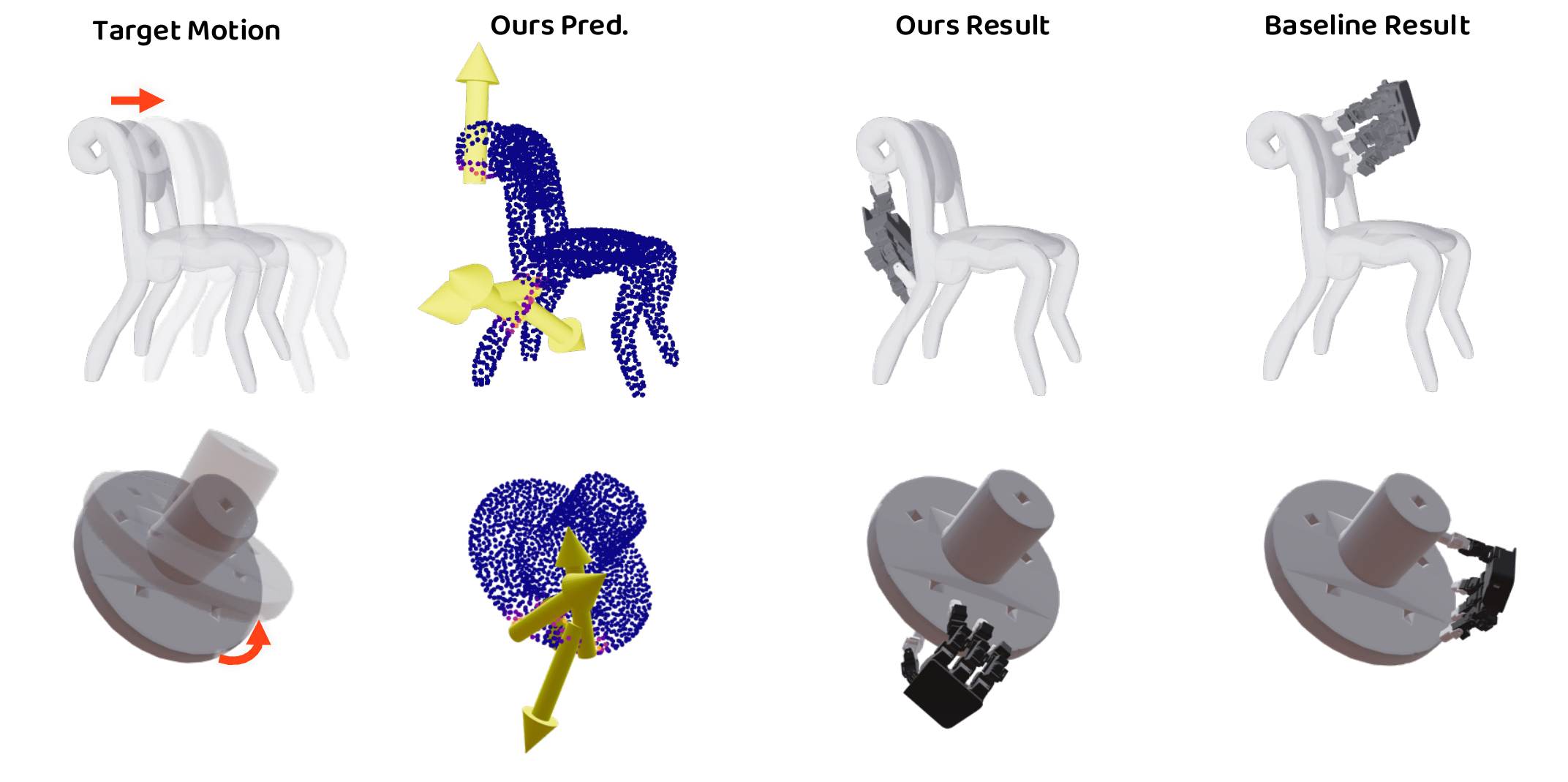}
\vspace{-10pt}
\caption{Visualization of evaluation on rigid bodies}
\label{fig:rigid_vis}
\vspace{-10pt}
\end{figure}

The success of \textbf{Ours} and the failure of \textbf{Baseline} can be illustrated in Fig. \ref{fig:rigid_vis}. As for a big object (the first row in Fig. \ref{fig:rigid_vis}), \textbf{Ours} pushes it on the back to move it to the right, while as for a small object (the second row in Fig. \ref{fig:rigid_vis}), \textbf{Ours} grasps its edge to lift it on one side. However, without \textbf{WBR}, \textbf{Baseline} can provide neither a contact solution to directly achieve the target motion, nor a force-closure grasp. 

\subsection{Generalization over Novel Robot Hands}
Our network is trained on multiple robot hands, we evaluate if our model has the ability to generalize to novel robot hands, such as Allegro. In other words, we have not trained our model with the data annotated by Allegro hand. Using the Allegro hand, \textbf{Ours} attained an average accuracy of 94.0\% on the test set, compared to 18.7\% for the \textbf{Baseline} (described in ), demonstrating effective generalization to novel robotic hands.
So the generalization performance is shown in Tab. \ref{tab:q7q8} to answer Q8. The challenge of the wrench-based task on Allegro is reflected by the results of \textbf{Baseline}. However, It is surprising to see that \textbf{Ours} achieves a competitive performance, proving that our network does have the potential of novel robot hands generalization.

\subsection{Real World Experiments}
\begin{figure}[H]
\vspace{-10pt}
\centering
\includegraphics[width=0.48\textwidth]{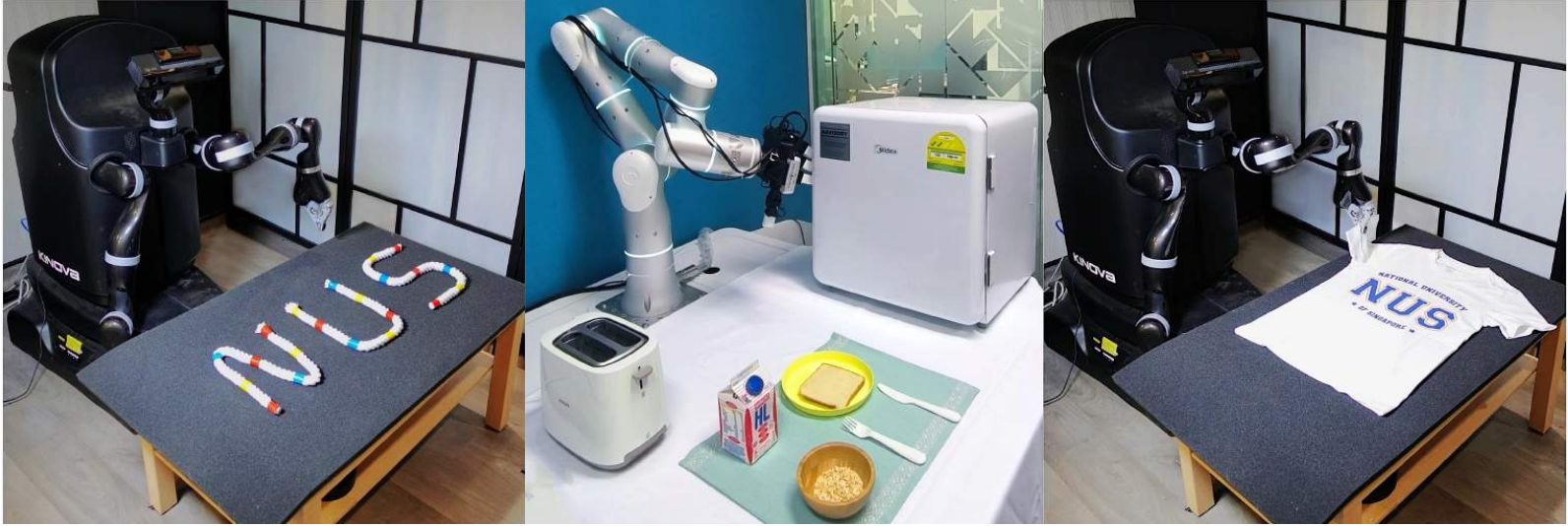}
\vspace{-10pt}
\caption{Realworld experiment settings.}
\label{fig:realworld}
\vspace{-10pt}
\end{figure}
\begin{figure*}[t]
\vspace{-10pt}
\centering
\includegraphics[width=0.98\textwidth]{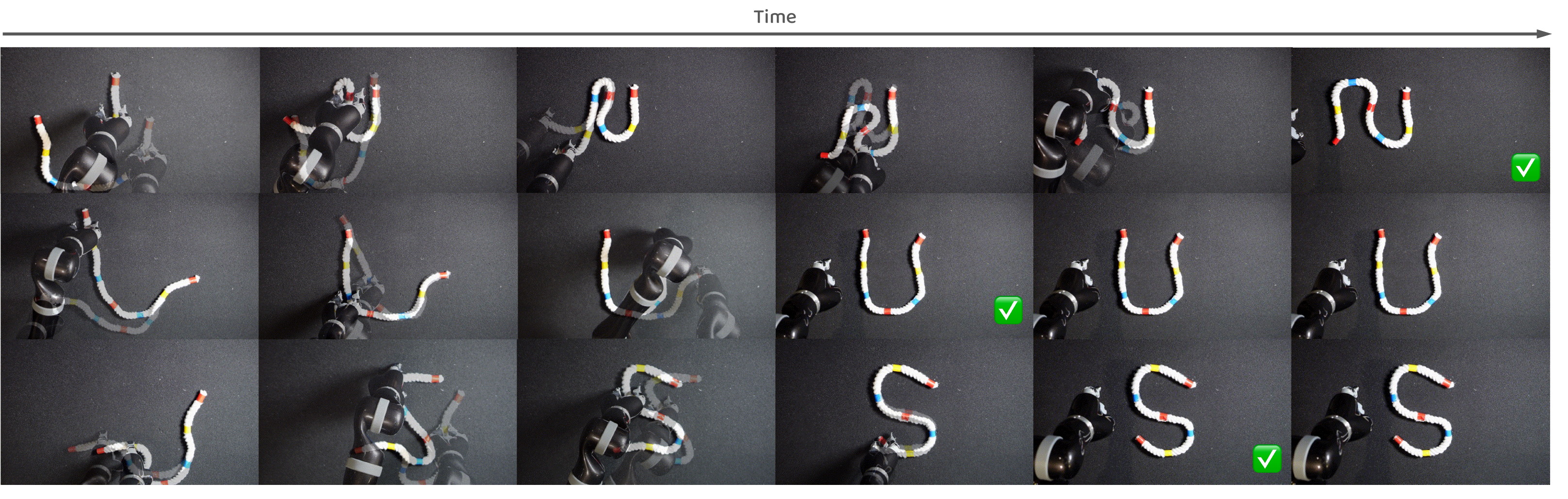}
\vspace{-10pt}
\caption{Test Cases of Rope Rearrangement.}
\label{fig:ropedemo}
\vspace{-10pt}
\end{figure*}

\begin{table}[t]
\centering
\begin{tabular}{@{}cccc@{}}
\toprule
Letter & Mean Dis(m) & Max Dis(m) & Success Rate \\ \midrule
N      & 0.0210      & 0.0376     & 10/10        \\
U      & 0.0186      & 0.0394     & 9/10         \\
S      & 0.0183      & 0.0384     & 8/10         \\ \bottomrule
\end{tabular}
\caption{Rope Rearrangement Experiment Results.}
\label{tab:rope}
\vspace{-10pt}
\end{table}

In this section, we evaluate our system's performance of manipulating various rigid, articulated rigid and deformable objects in several real-world settings. 

\textbf{Rope Rearrangement} We design a rope rearrangement experiment for a Kinova MOVO robot arm to rearrange a random reset rope to letters ``NUS''.  We perform 10 times for each letter in ``NUS''. Specifically, we take RGBD images for both the  current scene and the goal scene, and get the rope point cloud with SAM \cite{kirillov2023segany}. We then approximate the rope point clouds with several cylinders so that we can calculate the per-point target motion by obtaining the cylinder's planar rotation and translation from current to goal scene. We iteratively grasp the point with the highest predicted contact heatmap value and move to the predicted target point. A test case visualization is shown in Fig. \ref{fig:ropedemo}. As a result, our model achieves 90\% success rate, as shown in Tab. \ref{tab:rope}, indicating that our model is accurate in generating grasp points for deformation objects like ropes. 
% This action will be executed repeatedly until all points in the rope point cloud have a max $L2$ error $<0.05m$ and a mean $L2$ error $<0.03m$ w.r.t. its target point in this stage. 

\textbf{Breakfast Preparation}
We design a complex breakfast preparation experiment for a single Flexiv robot arm and a LeapHand as the manipulator. In this setting, the robot will open the fridge door, take out the milk box, place the milk box on the table, pick up a piece of bread, put the bread into the toaster, open the toaster, take out the cooked bread and place it to the plate. Among these objects, the fridge door is an articulated object, while the milk box is a normal rigid body object, and the bread is treated as a 3D deformation object. 

\textbf{Cloth Folding} We design a cloth folding experiment for a Kinova MOVO robot arm to fold a T-shirt. We extract the T-shirt point clouds using SAM~\cite{kirillov2023segany}. We design 3 substages of clothes folding and obtain the per-point target motion in DiffCloth\cite{li2022diffcloth}. Movo will grasp the point with the highest heatmap value and move to the predicted target point.

\end{document}